\documentclass[pdflatex,sn-apa,iicol]{sn-jnl}
\PassOptionsToPackage{numbers, authoryear}{natbib}

\jyear{2021}%

\theoremstyle{thmstyleone}%
\theoremstyle{thmstyletwo}%

\theoremstyle{thmstylethree}%
\usepackage{amsfonts}
\usepackage{amsmath}
\usepackage{amssymb}
\DeclareMathOperator*{\argmax}{arg\,max}
\DeclareMathOperator*{\argmin}{arg\,min}
\DeclareGraphicsExtensions{.pdf,.eps,.png,.jpg,.tif,.tiff,.ps}
\graphicspath{{.}{figs/}}

\usepackage{amsthm}
\usepackage{comment}
\usepackage[utf8]{inputenc} 
\usepackage[T1]{fontenc}    
\usepackage{hyperref}       
\usepackage{url}            
\usepackage{booktabs}       
\usepackage{amsfonts}       
\usepackage{nicefrac}       
\usepackage{microtype}
\usepackage{breakcites}
\usepackage{subcaption}
\usepackage{xspace}
\newcommand{\spiVAE}{$\pi$VAE\xspace}
\newcommand{\Matern}{Mat\'ern\xspace}
\newcommand{\spiVAEs}{$\pi$VAEs\xspace}
\newcommand{\squishlisttwo}{
 \begin{list}{$\bullet$}
  { \setlength{\itemsep}{0pt}
    \setlength{\parsep}{0pt}
    \setlength{\topsep}{0pt}
    \setlength{\partopsep}{0pt}
    \setlength{\leftmargin}{1.5em}
    \setlength{\labelwidth}{1.5em}
    \setlength{\labelsep}{0.5em} } }
\newcommand{\squishend}{
  \end{list}  }
\usepackage{placeins}
\begin{document}

\title[\spiVAE]{\spiVAE: a stochastic process prior for Bayesian deep learning with MCMC}

\author*[1,2]{\fnm{Swapnil} \sur{Mishra}}\email{s.mishra@imperial.ac.uk}
\equalcont{These authors contributed equally to this work.}
\author[3]{\fnm{Seth} \sur{Flaxman}}
\equalcont{These authors contributed equally to this work.}
\author[4]{\fnm{Tresnia} \sur{Berah}}
\author[4]{\fnm{Harrison} \sur{Zhu}}
\author[4]{\fnm{Mikko} \sur{Pakkanen}}
\author[1,2]{\fnm{Samir} \sur{Bhatt}}
\equalcont{These authors contributed equally to this work.}

\affil*[1]{\orgdiv{MRC Centre for Global Infectious Disease Analysis, Jameel Institute for Disease and Emergency Analytics, School of Public Health}, \orgname{Imperial College London}, \orgaddress{ \city{London}, \country{UK}}}

\affil[2]{\orgdiv{Section of Epidemiology, Department of Public Health}, \orgname{University of Copenhagen}, \orgaddress{ \city{Copenhagen}, \country{Denmark}}}

\affil[3]{\orgdiv{Department of Computer Science}, \orgname{University of Oxford}, \orgaddress{ \city{Oxford}, \country{UK}}}

\affil[4]{\orgdiv{Department of Mathematics}, \orgname{Imperial College London}, \orgaddress{ \city{London}, \country{UK}}}


\abstract{Stochastic processes provide a mathematically elegant way to model complex data. In theory, they provide flexible priors over function classes that can encode a wide range of interesting assumptions. However, in practice efficient inference by optimisation or marginalisation is difficult, a problem further exacerbated with big data and high dimensional input spaces. 
We propose a novel variational autoencoder (VAE) called the prior encoding variational autoencoder (\spiVAE). \spiVAE is a new continuous stochastic process. We use \spiVAE to learn low dimensional embeddings of function classes by combining a trainable feature mapping with generative model using a VAE. We show that our framework can accurately learn expressive function classes such as Gaussian processes, but also properties of functions such as their integrals. For popular tasks, such as spatial interpolation, \spiVAE achieves state-of-the-art performance both in terms of accuracy and computational efficiency. Perhaps most usefully, we demonstrate an elegant and scalable means of performing fully Bayesian inference for stochastic processes within probabilistic programming languages such as Stan.}

\keywords{Bayesian inference, MCMC, VAE, Spatio-temporal}

\maketitle

\section{Introduction}
A central task in machine learning is to specify a function or set of functions that best generalises to new data. Stochastic processes \citep{ross1996,pavliotis2014stochastic} provide a mathematically elegant way to define a class of functions, where each element from a stochastic process is a (usually infinite) collection of random variables. Popular examples of stochastic processes in computational statistics and machine learning are Gaussian processes \citep{Rasmussen2006}, Dirichlet processes \citep{antoniak1974}, log-Gaussian Cox processes \citep{Moller_LGCP}, Hawkes processes \citep{Hawkes1971}, Mondrian processes \citep{Roy2009} and Gauss-Markov processes \citep{Lindgren2011}. Many of these processes are intimately connected with popular techniques in deep learning, for example, both the infinite width limit of a single layer neural network and the evolution of a deep neural network by gradient descent are Gaussian processes \citep{Neal1996,Jacot2018}. However, while stochastic processes have many favourable properties, they are often cumbersome to work with in practice. For example, inference and prediction using a Gaussian process requires matrix inversions that scale cubicly with data size, log-Gaussian Cox processes require the evaluation of an intractable integral and Markov processes are often highly correlated. Bayesian inference can be even more challenging due to complex high dimensional posterior topologies. Gold standard evaluation of posterior expectations is done by Markov Chain Monte Carlo (MCMC) sampling, but high auto-correlation, narrow typical sets \citep{Betancourt2017} and poor scalability have prevented use in big data and complex model settings. A plethora of approximation algorithms exist \citep{Minka2001,Ritter2018,Lakshminarayanan2017,Welling2011,Blundell2015}, but few actually yield accurate posterior estimates \citep{Yao2018,huggins2019,hoffman2013stochastic,yao2019quality}. In this paper, rather than relying on approximate Bayesian inference to solve complex models, we extend variational autoencoders (VAE)~\citep{Kingma2014b,rezende2014stochastic} to develop portable models that can work with state-of-the-art Bayesian MCMC software such as Stan \citep{Carpenter2017a}. Inference on the resulting models is tractable and yields accurate posterior expectations and uncertainty.

 An autoencoder~\citep{hinton2006reducing} is a model comprised of two component networks. The encoder $e : \mathcal{X} \rightarrow \mathcal{Z}$ encodes inputs from space $\mathcal{X}$  into a latent space $\mathcal{Z}$ of lower dimension than $\mathcal{X}$. The decoder $d : \mathcal{Z} \rightarrow \mathcal{X}$ decodes latent codes in $\mathcal{Z}$ to reconstruct the input. The parameters of $e$ and $d$ are learned through the minimisation of a reconstruction loss on a training dataset. A VAE extends the autoencoder into a generative model~\citep{Kingma2014b}. In a VAE, the latent space $\mathcal{Z}$ is given a distribution, such as standard normal, and a variational approximation to the posterior is estimated. In a variety of applications, VAEs do a superb job reconstructing training datasets and enable the generation of new data: samples from the latent space are decoded to generate synthetic data \citep{kingma2019introduction}. In this paper we propose a novel use of VAEs: we learn low-dimensional representations of samples from a given function class (e.g.~sample paths from a Gaussian process prior). We then use the resulting low dimensional representation and the decoder to perform Bayesian inference. 
 
  One key benefit of this approach is that we decouple the prior from inference to encode arbitrarily complex prior function classes, without needing to calculate any data likelihoods. A second key benefit is that when inference is performed, our sampler operates in a low dimensional, uncorrelated latent space which greatly aids efficiency and computation, as demonstrated in the spatial statistics setting in PriorVAE~\citep{semenova2022prior}. One limitation of this approach (and of PriorVAE) is that we are restricted to encoding finite-dimensional priors, because VAEs are not stochastic processes. To overcome this limitation, we take as inspiration the Karhunen-Loève decomposition of a stochastic process as a random linear combination of basis functions and introduce a new VAE called the prior encoding VAE (\spiVAE). \spiVAE is a valid stochastic process by construction, it is capable of learning a set of basis functions, and it incorporates a VAE, enabling simulation and highly effective fully Bayesian inference. 
 
  
  We employ a two step approach: first, we encode the prior using our novel architecture; second we use the learnt basis and decoder network---a new stochastic process in its own right---as a prior, combining it with a likelihood in a fully Bayesian modeling framework, and use MCMC to fit our model and infer the posterior. 
  We believe our framework's novel decoupling into two stages  is critically important for many complex scenarios, because we do not need to compromise in terms of either the expressiveness of deep learning or accurately characterizing the posterior using fully Bayesian inference.
   
  We thus avoid some of the drawbacks of other Bayesian deep learning approaches which rely solely on variational inference, and the drawbacks of standard MCMC methods for stochastic processes which are inefficient and suffer from poor convergence. 

  Taken together, our work is an important advance in the field of Bayesian deep learning, providing a practical framework combining the expressive capability of deep neural networks to encode stochastic processes with the effectiveness of fully Bayesian and highly efficient gradient-based MCMC inference to fit to data while fully characterizing uncertainty.  
  
  Once a \spiVAE is trained and defined, the complexity of the decoder scales linearly in the size of the largest hidden layer. Additionally, because the latent variables are penalised via the KL term from deviating from a standard normal distribution, the latent space is approximately uncorrelated, leading to high effective sample sizes in MCMC sampling. 
  The main contributions of this paper are:
  \squishlisttwo
      \item We apply the generative framework of VAEs to perform full Bayesian inference. We first encode priors in training and then, given new data, perform inference on the latent representation while keeping the trained decoder fixed.
      \item We propose a new generative model, \spiVAE, that  generalizes VAEs to be able to learn priors over both functions and properties of functions. We show that \spiVAE is a valid (and novel) stochastic process by construction.
      \item We show the performance of \spiVAE on a range of simulated and real data, and show that \spiVAE achieves state-of-the-art performance in a spatial interpolation task. 
  \squishend
  The  rest  of  this  paper  is  structured  as  follows. Section~\ref{sec:methods} details the proposed framework and the generative model along with toy fitting examples. The experiments on large real world datasets  are outlined in Section~\ref{sec:results}.  We discuss our findings and conclude in Section~\ref{sec:discussion}.

\section{Methods}
\label{sec:methods}
\subsection{Variational Autoencoders (VAEs)}
\label{subsec-simple-vae}
A standard VAE has three components:
\begin{enumerate}
    \item 
an encoder  network $e(x,\gamma)$ which encodes inputs $x \in \mathcal{X}$ using learnable parameters $\gamma$,
\item random variables $z$ for the latent subspace,
\item a decoder  network $d(z,\psi)$ which decodes latent embeddings $z$ using learnable parameters $\psi$.
\end{enumerate}
In the simplest case we are given inputs $x\in \mathbb{R}^d = \mathcal{X}$ such as a flattened image or discrete time series. The encoder $e(x,\gamma)$ and decoder $d(z,\psi)$ are fully connected neural networks (though they could include convolution or recurrent layer). The output of the encoder network are vectors of mean and standard deviation parameters $z_{\mu}$ and $z_{sd}$. These vectors can thus be used to define the random variable $\mathcal{Z}$ for the latent space:
\begin{align}
\label{eq:VAE}
[z_\mu,z_{sd}]^{\top} &= e(x,\gamma)  \\
\mathcal{Z} &\sim \mathcal{N}(z_\mu,z_{sd}^{2}\mathbb{I})
\end{align}
For random variable $\mathcal{Z}$, the decoder network reconstructs the input by producing $\hat x$:
\begin{equation}
\label{eq:VAE1}
 \hat{x} = d(\mathcal{Z},\psi)
 \end{equation}




To train a VAE, a variational approximation is used to estimate the posterior distribution $$p(\mathcal{Z} \mid x,\gamma,\psi) \propto  p(x \mid \mathcal{Z},\gamma,\psi) \times p(\mathcal{Z})$$ The variational approximation greatly simplifies inference by turning a marginalisation problem into an optimisation problem. Following \citep{Kingma2014}, the optimal parameters for the encoder and decoder are found by maximising the evidence lower bound:  
\begin{align}
\argmax_{\gamma,\psi} ~~ &\mathbb{E}_\mathcal{Z}\bigg[\log p\left(x\mid\mathcal{Z},\gamma,\psi\right) \nonumber\\
&- \text{KL}\left(\mathcal{Z} ~\|~ \mathcal{N}(0,\mathbb{I})\right)\bigg]\label{eq:VAE_loss}
\end{align}
The first term in Eq.~\eqref{eq:VAE_loss} is the likelihood quantifying how well $\hat{x}$ matches $x$. In practice we can simply adopt the mean squared error loss directly, referred to as the reconstruction loss, without taking a probabilistic perspective. The second term is a Kullback-Leibler divergence  to ensure that $\mathcal{Z}$ is as similar as possible to the prior distribution, a standard normal. Again, this second term can be specified directly without the evidence lower bound derivation: we view the KL-divergence as a regularization penalty to ensure that the latent parameters are approximately uncorrelated by penalizing how far they deviate from $\mathcal{N}(0,\mathbb{I})$. 

Once training is complete, we fix $\psi$, and use the decoder as a generative model. To simplify subsequent notation we refer to a fully trained decoder as $d(z)$.  Generating a new sample is simple: first draw a random variable $\mathcal{Z} \sim \mathcal{N}(0,\mathbb{I})$ and then apply the decoder, which is a deterministic transformation to obtain $d(\mathcal{Z})$. We see immediately that $d(\mathcal{Z})$ is itself a random variable. In the next section, we will use this generative model as a prior in a Bayesian framework by linking it to a likelihood to obtain a posterior.

\subsection{VAEs for Bayesian inference}
 VAEs have been typically used in the literature to create or learn a generative model of observed data  \citep{Kingma2014b}, such as images. \citep{semenova2022prior} introduced a novel application of VAEs in a Bayesian inference setting, using a two stage approach that is closely related to ours. In brief, in the first stage, a VAE is trained to encode and decode a large dataset of vectors consisting of samples drawn from a specified prior $p(\theta)$ over random vectors. In the second stage, the original prior is replaced with the approximate prior: 
 $\theta := d(\mathcal{Z})$ where  $\mathcal{Z} \sim \mathcal{N}(0,\mathbb{I})$.
 
To see how this works in a Bayesian inference setting, consider a likelihood $p(y\mid \theta)$ linking the parameter $\theta$ to data $y$. Bayes' rule gives the unnormalized posterior:
\begin{equation}
p(\theta \mid y) \propto p(y \mid \theta) \times  p(\theta)
\end{equation}
The trained decoder serves as a drop-in replacement for the original prior class in a Bayesian setting:
 \begin{align}
 p(\mathcal{Z}\mid y,d) & \propto p(y\mid d(\mathcal{Z})) \times 
 p(\mathcal{Z})\,.  \label{eqn:VAE_bayes}
 \end{align}





The implementation within a probabilistic programming language is very straightforward: a standard normal prior and deterministic function (the decoder) are all that is needed.

It is useful to contrast the inference task from Eq.~\eqref{eqn:VAE_bayes} to a Bayesian neural network (BNN) \citep{Neal1996} or Gaussian process in primal form \citep{Rahimi2007}. In a BNN with parameters $\omega$ and  hyperparameters $\lambda$, the unnormalised posterior would be 
\begin{equation}
  \label{eqn:BNN}
    p(\omega,\lambda\mid y) \propto p(y\mid \omega,\lambda) \times p(\omega\mid\lambda) \times p(\lambda)\,.
\end{equation}
The key difference between Eq.~\eqref{eqn:BNN} and Eq.~\eqref{eqn:VAE_bayes} is the term $p(\omega\mid\lambda)$. The dimension of $\omega$ is typically huge, sometimes in the millions, and is conditional on $\lambda$, whereas in Eq.~\eqref{eqn:VAE_bayes} the latent dimension of $\mathcal{Z}$ is typically small ($<50$), uncorrelated and unconditioned. Full batch MCMC training is typically prohibitive for BNNs due to large datasets and the high-dimensionality of $\omega$, but approximate Bayesian inference algorithms tend to poorly capture the complex posterior \citep{yao2019quality,Yao2018}. Additionally, $\omega$ tends to be highly correlated, making efficient MCMC nearly impossible. Finally, as the dimension and depth increases, the posterior distribution suffers from complex multimodality, and concentration to a narrow typical set ~\citep{Betancourt2017}. By contrast, off-the-shelf MCMC methods are very effective for equation \eqref{eqn:VAE_bayes}  because the prior space they need to explore is as simple as it could be: a standard normal distribution, while the complexity of the model lives within the deterministic (and differentiable) decoder. In a challenging spatial statistics setting,  \citep{semenova2022prior} used this approach and achieved MCMC effective sample sizes {\em exceeding} actual sample sizes, due to the incredible efficiency of the MCMC sampler.

\begin{figure*}[!tbp]
  \centering
  \begin{subfigure} {.48\textwidth}
  \includegraphics[width=\textwidth,height=0.14\textheight]{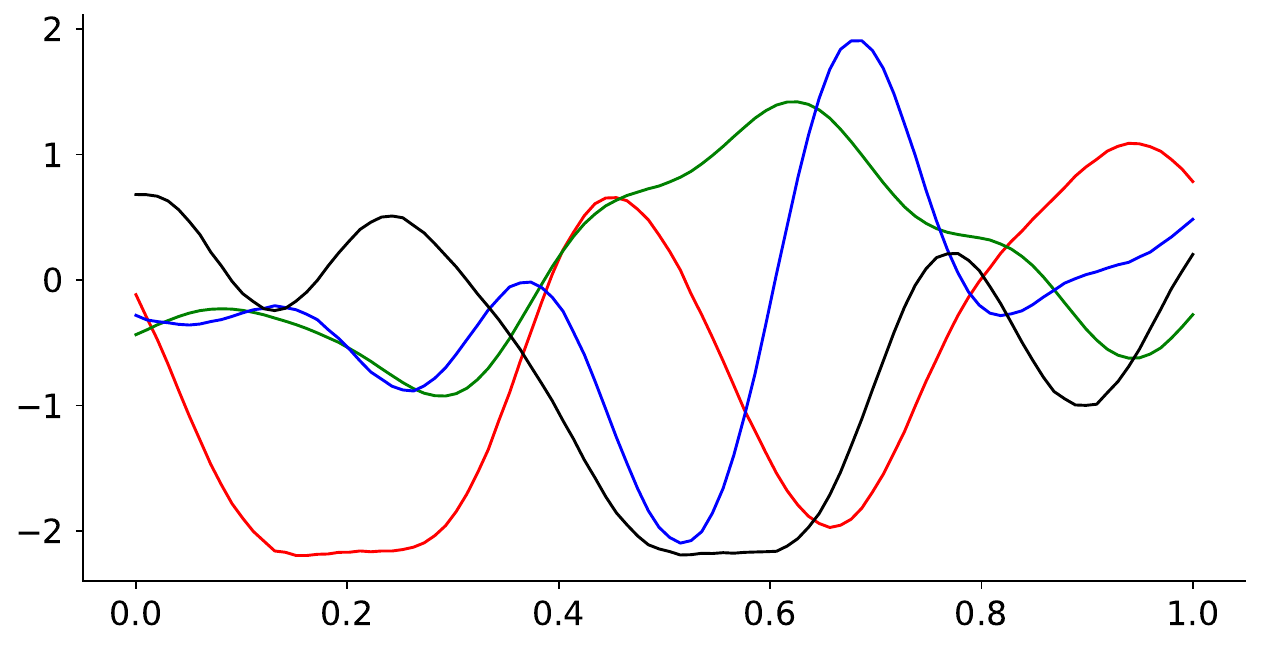}
  \caption{}
\end{subfigure}
\begin{subfigure} {.48\textwidth}
  \includegraphics[width=\textwidth,height=0.14\textheight]{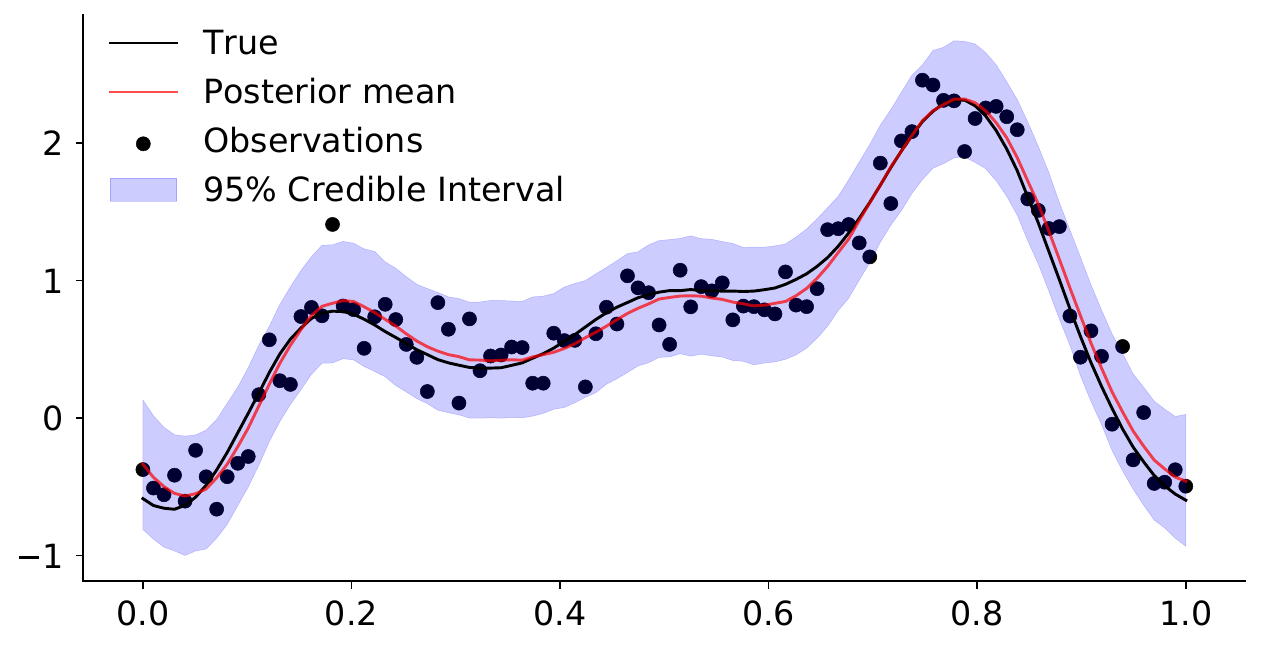}
\caption{}
\end{subfigure}
   \caption{
Learning functions with VAE: (a) Prior samples from a VAE trained on Gaussian process samples (b) we fit our VAE model to data drawn from a GP (blue) plus noise (black points). The  posterior mean of our model is in red with the $95\%$ epistemic credible intervals shown in purple.
  }
  \label{fig:1}
\end{figure*}

An example of using VAEs to perform inference is shown in Figure~\ref{fig:1} where we train a VAE with latent dimensionality 10 on samples drawn from a zero mean Gaussian process with RBF kernel ($K(\delta)=e^{- \delta^2/8^2}$) observed on the grid $0, 0.01, 0.02, \ldots, 1.0$. In Figure~\ref{fig:1}  we closely recover the true function and correctly estimate the data noise parameter. Our MCMC samples showed virtually no autocorrelation, and all diagnostic checks were excellent (see Appendix). Solving the equivalent problem using a Gaussian process prior would not only be considerably more expensive ($\mathcal{O}(n^3)$) but correlations in the parameter space would complicate MCMC sampling and necessitate very long chains to achieve even modest effective sample sizes.

This example demonstrates the promise that VAEs hold to improve Bayesian inference by encoding  function classes in a two stage process. While this simple example proved useful in some settings~\citep{semenova2022prior}, inference and prediction is not  possible at new input locations, because a VAE is not a  stochastic process. As described above, a VAE provides a novel prior over random vectors. Below, we take the next step by introducing  \spiVAE, a new stochastic process capable of approximating useful and widely used priors over function classes, such as Gaussian processes.

\subsection{Encoding stochastic processes with \spiVAE}
\label{subsec:sp-vae}

To create a model with the ability to perform inference on a wide range of problems we have to ensure that it is a valid stochastic process. Previous attempts in deep learning in this direction have been inspired by the Kolmogorov Extension Theorem and have focused on extending from a finite-dimensional distribution to a stochastic process. Specifically,~\citep{Garnelo2018} introduced an aggregation step (typically an average) to create an order invariate global distribution. However, as noted by \citep{AttentiveNP}, this can lead to underfitting. 

We take a different approach with \spiVAE, inspired by the Karhunen-Loève Expansion~\citep{karhunen1947linear, loeve1948functions}. Recall that a centered stochastic process $f(s)$ can be written as an infinite sum:
\begin{equation}
\label{eq:KL}
f(s) = \sum_{j=1}^\infty \beta_j \phi_j(s)
\end{equation}
for pairwise uncorrelated random variables $\beta_j$ and  continuous real-valued functions forming an orthonormal basis $\phi_j(s)$. The random $\beta_j$'s provide a linear combination of a fixed set of basis functions, $\phi_j$. This perspective has a long history in neural networks, cf.~radial basis function networks.

What if we consider a trainable, deep learning parameterization of Eq.~\eqref{eq:KL} as inspiration? We need to learn deterministic basis functions while allowing the $\beta_j$'s to be random. Let $\Phi(s)$ be a feature mapping with weights $w$, i.e.~a feed-forward  neural network architecture over the input space, representing the basis functions. Let $\beta$ be a vector of weights on the basis functions, so $f(s) = \beta^{\top} \Phi(s)$. We use a VAE architecture to encode and decode $\beta$, meaning we maintain the random variable perspective and at the same time learn a flexible low-dimensional non-linear generative model. 

How can we specify and train this model? As with the VAE in the previous section, \spiVAE is trained on draws from a prior. Our goal is to encode a stochastic process prior $\Pi$, so we consider $i = 1, \ldots, N$ function realizations denoted $f_i(s)$. Each $f_i(s)$ is an infinite dimensional object, a function defined for all $s$, so we further assume that we are given a finite set of $K_i$ observation locations. We set $K_i = K$ for simplicity of implementation i.e.~the number of evaluations for each function is constant across all draws $i$. We denote the observed values as $y_i^k := f_i(s_i^k)$. The training dataset thus consists of $N$ sets of $K$ observation locations and function values: $$\{(s_i^1, y_i^1) \ldots, (s_i^K,y_i^K)\}_{i=1}^N$$ Note that the set of $K$ observation locations varies across the $N$ realizations. 

\begin{figure*}
  \centering
   \includegraphics[height=0.4\textheight]{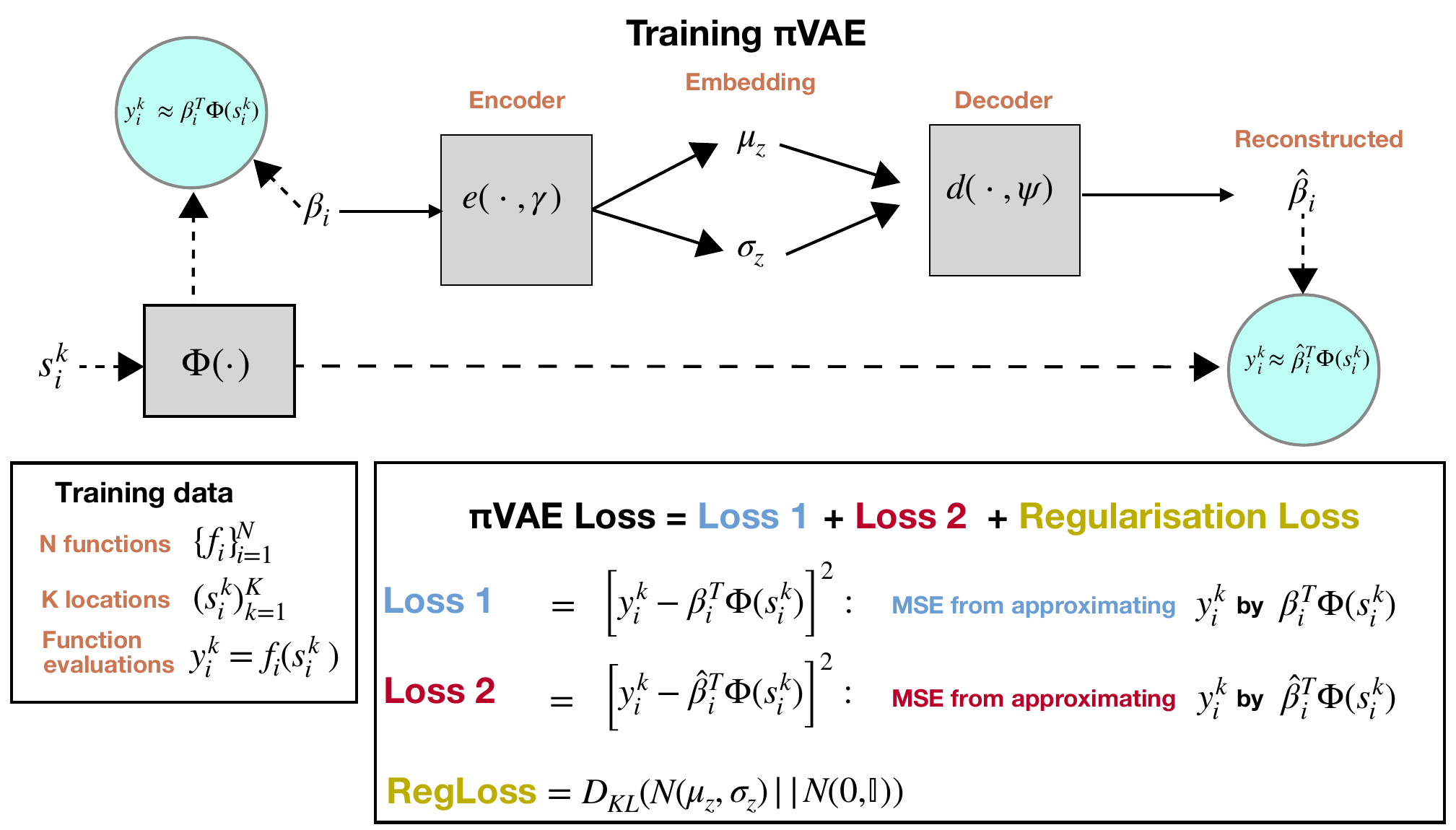}
  \caption{
Schematic description of end-to-end trainig procedure for \spiVAE including the reconstruction loss. Dashed arrows contribute to the loss, blue circles are reconstructions, and grey boxes are functions.}
  \label{fig:pi-vae-training}
\end{figure*}

We now return to the architecture of \spiVAE (Fig.~\ref{fig:pi-vae-training}). The feature mapping $\Phi(s)$ is shared across all $i = 1, \ldots, N$ function draws, so it consists of a feedforward neural network and is parameterized by a set of global parameters $w$ which must be learned. However, a particular random realization $f_i(s)$ is represented by a random vector $\beta_i$, for which we use a VAE architecture. We note the following non-standard setup: $\beta_i$ is a learnable parameter of our model, but it is also the input to the encoder of the VAE. The decoder attempts to reconstruct $\beta_i$ with an output $\hat \beta_i$. We denote the encoder and decoder as:
\begin{align}
[z_\mu,z_{sd}]^{\top} &= e(\beta,\gamma)  \\
\mathcal{Z} &\sim \mathcal{N}(z_\mu,z_{sd}^{2}\mathbb{I}) \\
 \hat{\beta} &= d(\mathcal{Z},\psi)
\end{align}

We are now ready to express the loss, which combines the two parts of the network, summing across all observations. Rather than deriving an evidence lower bound, we proceed directly to specify a loss function, in three parts. In the first, we use MSE to check the fit of the $\beta_i$'s and $\Phi$ to the data:
$$\mbox{Loss 1}: \frac{1}{N K} \sum_{i,k} (y_i^k - \beta_i^{\top}\Phi(s_i^k))^2$$
In the second, we use MSE to check the fit of the reconstructed $\hat \beta_i$'s and $\Phi$ to the data:
$$\mbox{Loss 2}: \frac{1}{N K} \sum_{i,k} (y_i^k - \hat \beta_i^{\top}\Phi(s_i^k))^2$$
We also require the standard variational loss:
$$ \text{KL}\left(\mathcal{Z}  ~\|~ \mathcal{N}(0,\mathbb{I})\right)$$
Note that we do not consider reconstruction loss $\|\beta_i - \hat \beta_i\|^2$ because in practice this did not improve training.

To provide more intuition: the feature map $\Phi(s)$ transforms each observed location to a fixed feature space  that is shared for all locations across all functions. $\Phi(s)$ could be an explicit feature representation for an RKHS (e.g.~an RBF network or a random Fourier feature basis \citep{Rahimi2007}), a neural network of arbitrary construction or, as we use in the examples in this paper, a combination of both. Following this transformation, a linear basis $\beta$ (which we obtain from a non-linear decoder network) is used to predict function evaluations at an arbitrary location. The intuition behind these two transformations is to learn the association between locations and observations while allowing for randomness---$\Phi$ provides the correlation structure over space and $\beta$ the randomness. Explicit choices can lead to existing stochastic processes: we can obtain a Gaussian process with kernel $k(\cdot,\cdot)$ using a single-layer linear VAE for $\beta$ (meaning the $\beta$s are simply standard normals) and setting $\Phi(s) = L^{\top} s$ for $L$ the Cholesky decomposition of the Gram matrix $K$ where $K_{ij} = k(s_i,s_j)$. 

In contrast to a standard VAE encoder that takes as input the data to be encoded, \spiVAE first transforms input data (locations) to a higher dimensional feature space via $\Phi$, and then connects this feature space to outputs, $y$, through a linear mapping, $\beta$. The \spiVAE decoder takes outputs from the encoder, and attempts to recreate $\beta$ from a lower dimensional probabilistic embedding. This re-creation, $\hat{\beta}$, is then used as a linear mapping with the \emph{same} $\Phi$ to get a reconstruction of the outputs $y$. It is crucial to note that a single global $\beta$ vector is \emph{not} learnt. Instead, for each function $i = 1, \ldots, N$ a $\beta_i$ is learnt. 

In terms of number of parameters, we need to learn $w$, $\gamma$, $\psi$,  $\beta_1, \ldots, \beta_N$. While this may seem like a huge computational task, $K$ is typically quite small ($<200$) and so learning can be relatively quick (dominated by matrix multiplication of hidden layers). 
Algorithm~\ref{algo:train} in the Appendix presents the step-by-step process of training \spiVAE.
 

\subsubsection{Simulation and Inference with \spiVAE}
Given a trained embedding $\Phi(\cdot)$ and trained decoder $d(z)$, 
we can use \spiVAE as a generative model to simulate sample paths $f$ as follows.
A single function $f$ is obtained by first drawing
$\mathcal{Z} \sim \mathcal{N}(0,\mathbb{I})$ and
defining $f(s) := d(\mathcal{Z})^{\top}\Phi(s)$. For a fixed $\mathcal{Z}$, $f(s)$ is a deterministic function---a sample path from \spiVAE defined for all $s$. Varying $\mathcal{Z}$ produces different sample paths. Computationally, $f$ can be efficiently evaluated at any arbitrary location $s$ using matrix algebra: $f(s) = d(\mathcal{Z})^{\top}\Phi(s)$. We remark that the stochastic process perspective is readily apparent: for a random variable $\mathcal{Z}$, $d(\mathcal{Z})^{\top}\Phi(s)$ is a random variable defined on the same probability space for all $s$. 

Algorithm~\ref{algo:sample} in the Appendix presents the step-by-step process for simulation with \spiVAE.

\spiVAE can be used for inference on new data pairs $(s_j,y_j)$, where the unnormalised posterior distribution is
 \begin{align}
  \label{eqn:spiVAE_bayes}
 p(\mathcal{Z}\mid d,y_j,s_j,\Phi) & \propto  p(y_j\mid d,s_j,\mathcal{Z},\Phi)p(\mathcal{Z})
 \end{align}
with likelihood $p(y_j\mid d,s_j,\mathcal{Z},\Phi)$ and prior $p(\mathcal{Z})$. MCMC can be used to efficiently obtain samples from the posterior distribution over $\mathcal{Z}$ using Equation~\eqref{eqn:spiVAE_bayes}.  An implementation in probabilistic programming languages such as Stan \citep{Carpenter2017a} is very straightforward.

The posterior predictive distribution of $y_j$ at a location $s_j$ is given by:
\begin{align}
\label{eqn:ppd}
&p(y_j \mid d, s_j, \Phi) =& \\ ~\nonumber
&\int p(y_j\mid d,s_j,\Phi, \mathcal{Z}) p(\mathcal{Z}\mid d,y_j,s_j,\Phi) d\mathcal{Z}  
 \end{align}

While equations Eqs.~\eqref{eqn:spiVAE_bayes}-\eqref{eqn:ppd} are written for a single location $s_j$, we can extend them to any arbitrary collection of locations without loss of generality, a necessary condition for  \spiVAE to be a valid stochastic process. Further, the distinguishing difference between Eq.~\eqref{eqn:VAE_bayes} and  Eqs.~\eqref{eqn:spiVAE_bayes}-\eqref{eqn:ppd} is conditioning on input locations and $\Phi$. It is $\Phi$ that ensures \spiVAE is a valid stochastic process. We formally prove this below.

Algorithm~\ref{algo:infer} in the Appendix presents the step-by-step process for  inference with \spiVAE.

 \subsubsection{\spiVAE is a stochastic process}
\vspace{-.15in}
~

{\bf Claim.} \spiVAE is a stochastic process. ~

Recall that, mathematically, a stochastic process is defined as a collection $\{f(s) : s \in S\}$, where $f(s)$ for each location $s \in S$ is a random variable on a common probability space $(\Omega,\mathcal{F},P)$, see, e.g., \citet[Definition 1.1]{pavliotis2014stochastic}. This technical requirement is necessary to ensure that for any locations $s_1,\ldots,s_n \in S$, the random variables $f(s_1),\ldots,f(s_n)$ have a well-defined joint distribution. Subsequently, it also ensures consistency. Namely, writing $f_i := f(s_i)$ and integrating $f_n$ out, we get
\begin{equation*}
p(f_1,\ldots,f_{n-1}) = \int_{f_n} p(f_1,\ldots,f_n) d f_n. 
\end{equation*}

{\bf Proof.} 
For \spiVAE, we have $f(\cdot) := d(\mathcal{Z}) \Phi(\cdot)$, where $\mathcal{Z}$ is a multivariate Gaussian random variable, hence  defined on some probability space $(\Omega,\mathcal{F},P)$. Since $d$ and $\Phi$ are deterministic (measurable) functions, it follows that $f(s_i) := d(\mathcal{Z}) \Phi(s_i)$ for any $i=1,\ldots,n$, is a random variable on $(\Omega,\mathcal{F},P)$, whereby $\{ f(s) : s \in S \}$ is a stochastic process. $\blacksquare$

We remark here that \spiVAE is a new stochastic process. If \spiVAE is trained on samples from a zero mean Gaussian process with a squared exponential covariance function, and similarly choose $\Phi$ to have the same covariance function, \emph{and} $d$ is linear, then \spiVAE will be a Gaussian process. But for a non-positive definite $\Phi$ and / or  non-linear $d$, even if \spiVAE is trained on samples from a Gaussian process, it will not truly be a Gaussian process, but some other stochastic process which approximates a Gaussian process. We do not know the theoretical conditions under which \spiVAE will perform better or worse than existing classes of stochastic processes; its general construction means that theoretical results will be challenging to prove in full generality. We demonstrate below that in practice, \spiVAE performs very well.

\subsection{Examples}
We first demonstrate the utility of our proposed \spiVAE model by fitting the simulated 1-D regression problem introduced in~\citep{hernandez2015probabilistic}. The training points for the dataset are created by uniform sampling of 20 inputs, $x$, between $(-4,4)$. The corresponding output is set as $y \sim \mathcal{N}(x^3, 9)$. We fit two different variants of \spiVAE, representing two different prior classes of functions. The first prior produces cubic monotonic functions and the second prior is a GP with an RBF kernel and a two layer neural network. We generated $10^4$ different function draws from both priors to train the respective \spiVAE. One important consideration in \spiVAE is to chose a sufficiently expressive $\Phi$, we used a RBF layer (see Appendix ~\ref{app:implementation}) with trainable centres coupled with two layer neural network with 20 hidden units each. We compare our results against 20,000 Hamiltonian Monte Carlo (HMC) samples~\citep{neal1993probabilistic} implemented using Stan~\citep{Carpenter2017a}. Details of the implementation for all the models can be found in the Appendix.
\begin{figure*}
  \centering
\begin{subfigure}{.32\textwidth}
   \includegraphics[width=\textwidth,height=0.16\textheight]{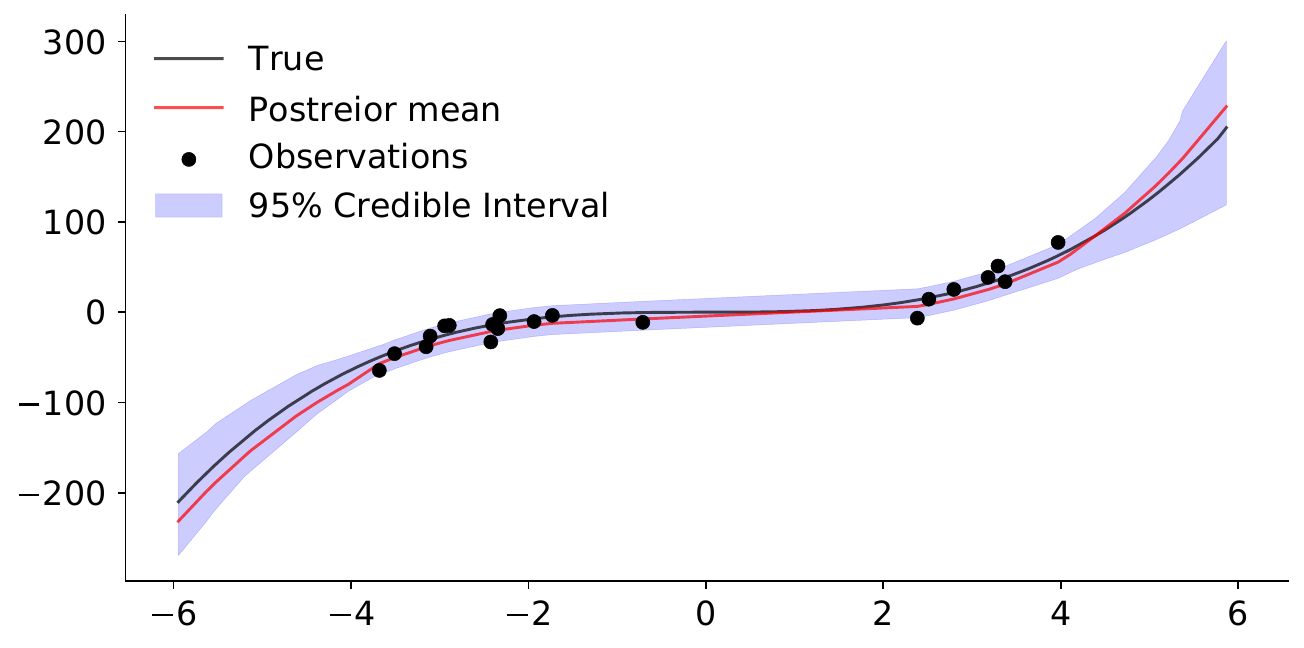}
   \caption{}
\end{subfigure}
 \begin{subfigure}{.32\textwidth}
  \includegraphics[width=\textwidth,height=0.16\textheight]{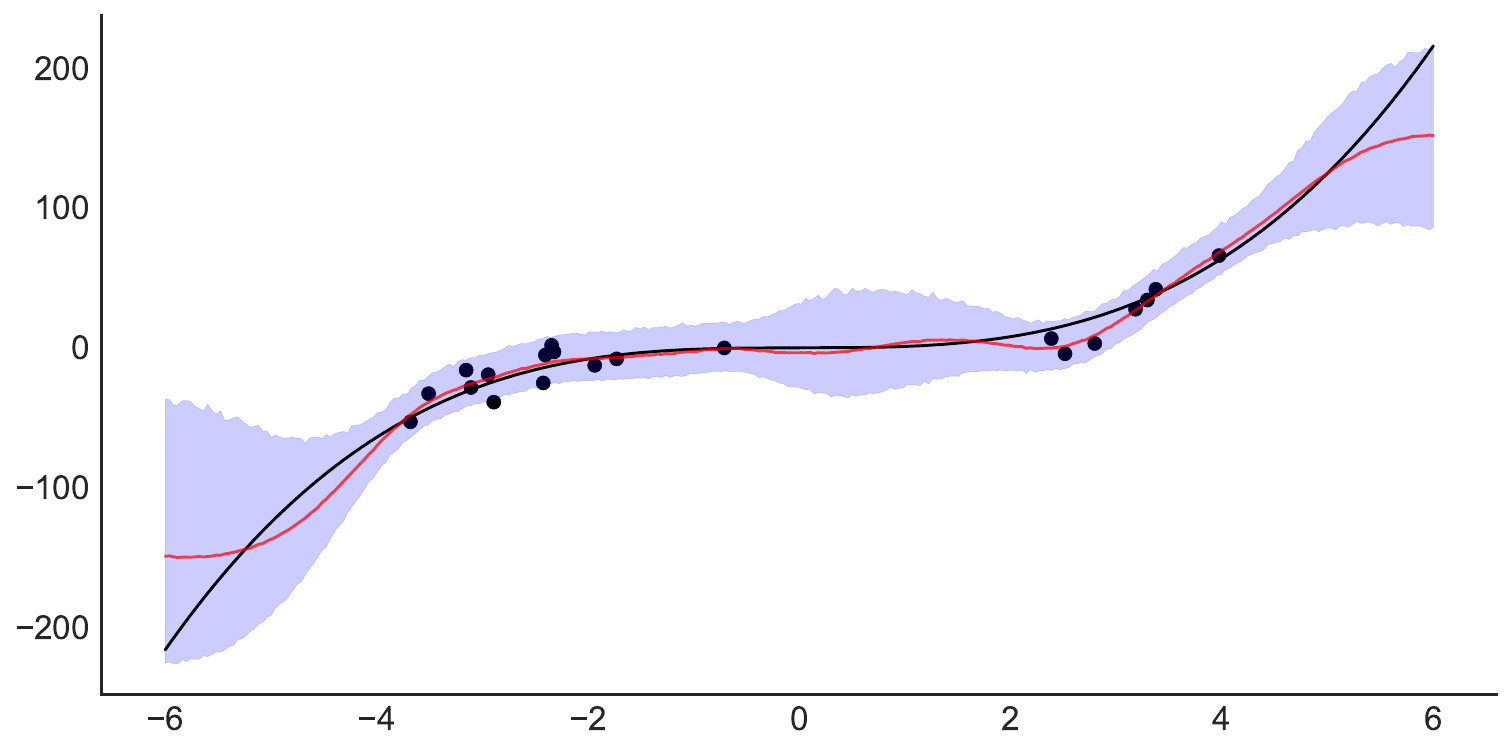}
  \caption{}
\end{subfigure}
\begin{subfigure}{.32\textwidth}
   \includegraphics[width=\textwidth,height=0.16\textheight]{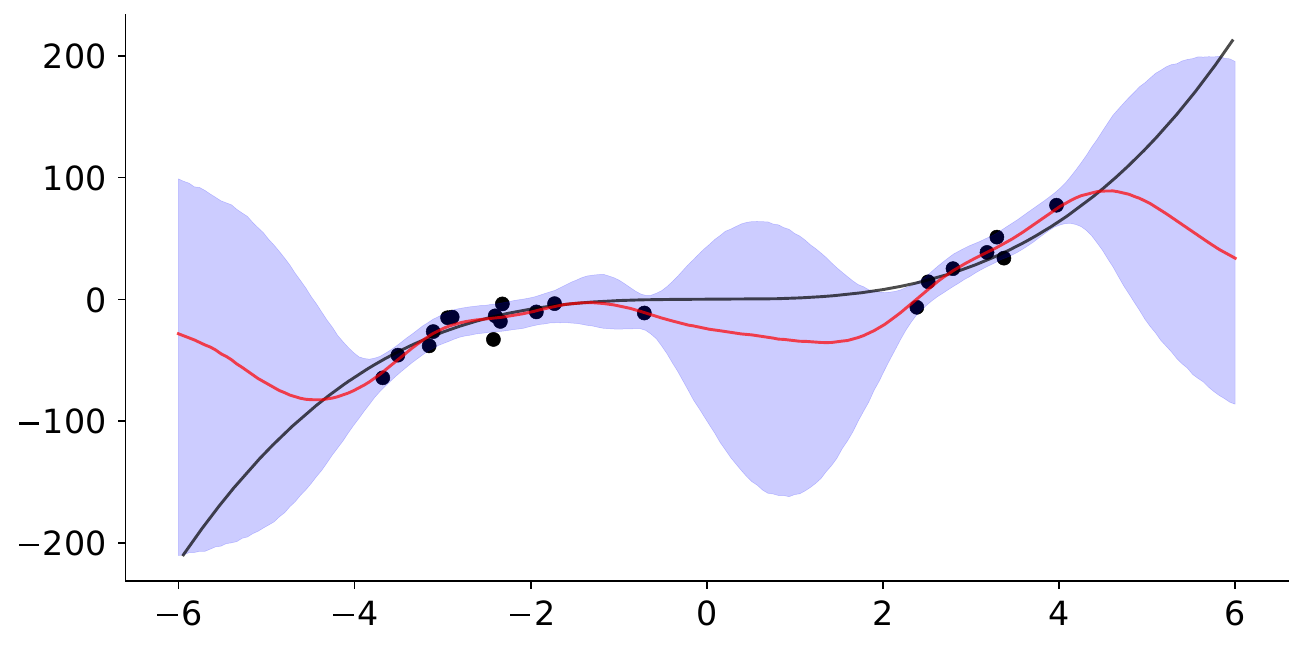}
   \caption{}
\end{subfigure}
  \caption{
Fitting to a cubic function with noise $y \sim \mathcal{N}(x^3, 9)$. (a) \spiVAE trained on a class of cubic functions, (b) \spiVAE trained on samples from a Gaussian process with RBF kernel and (c) is a Gaussian process with RBF kernel. All methods use Hamiltonian Markov Chain Monte Carlo for posterior inference.}
  \label{fig:1-d-uncertainty}
\end{figure*}
\begin{figure*}[t]
  \centering
  \begin{subfigure}{.48\textwidth}
  \includegraphics[width=\textwidth,height=0.18\textheight]{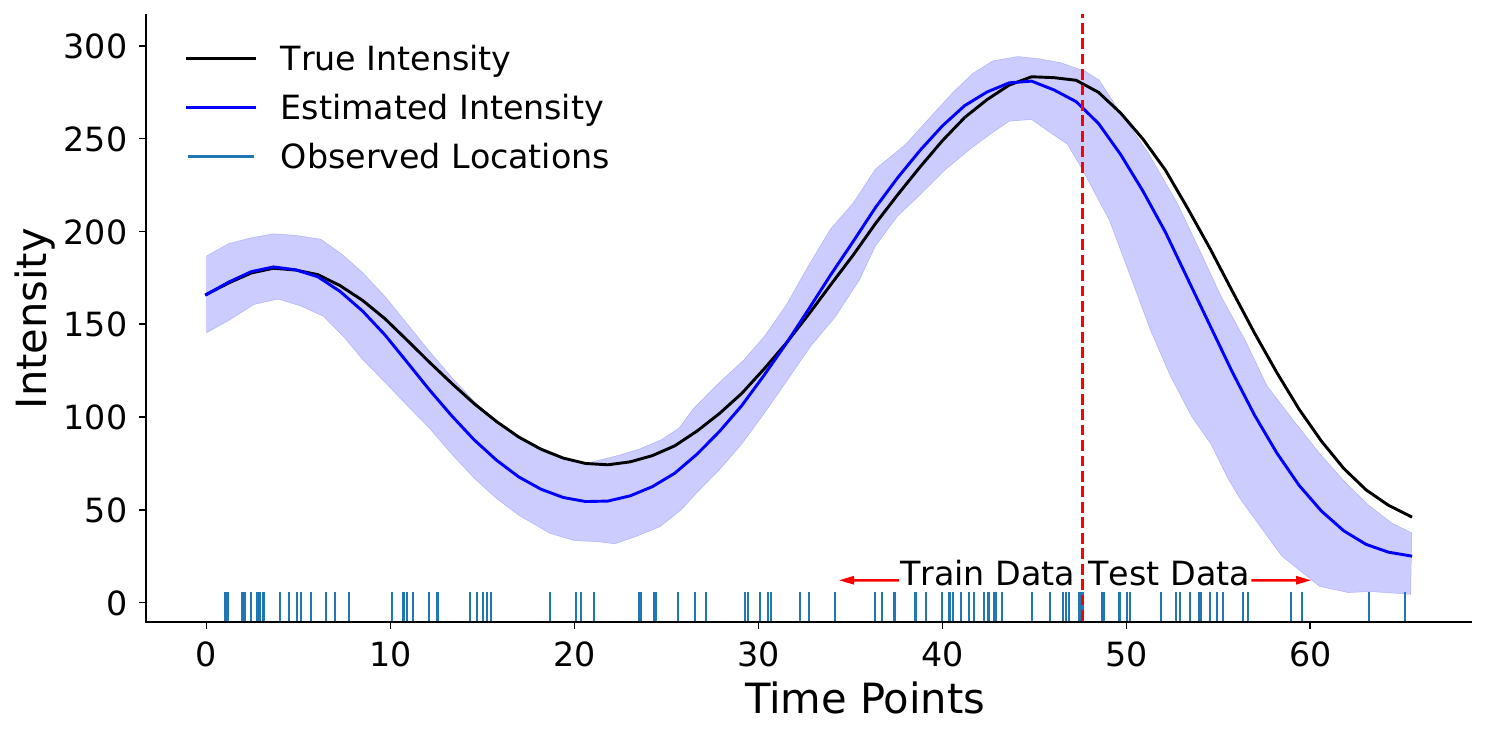}
  \caption{}
\end{subfigure} 
\begin{subfigure}{.48\textwidth}
  \includegraphics[width=\textwidth,height=0.18\textheight]{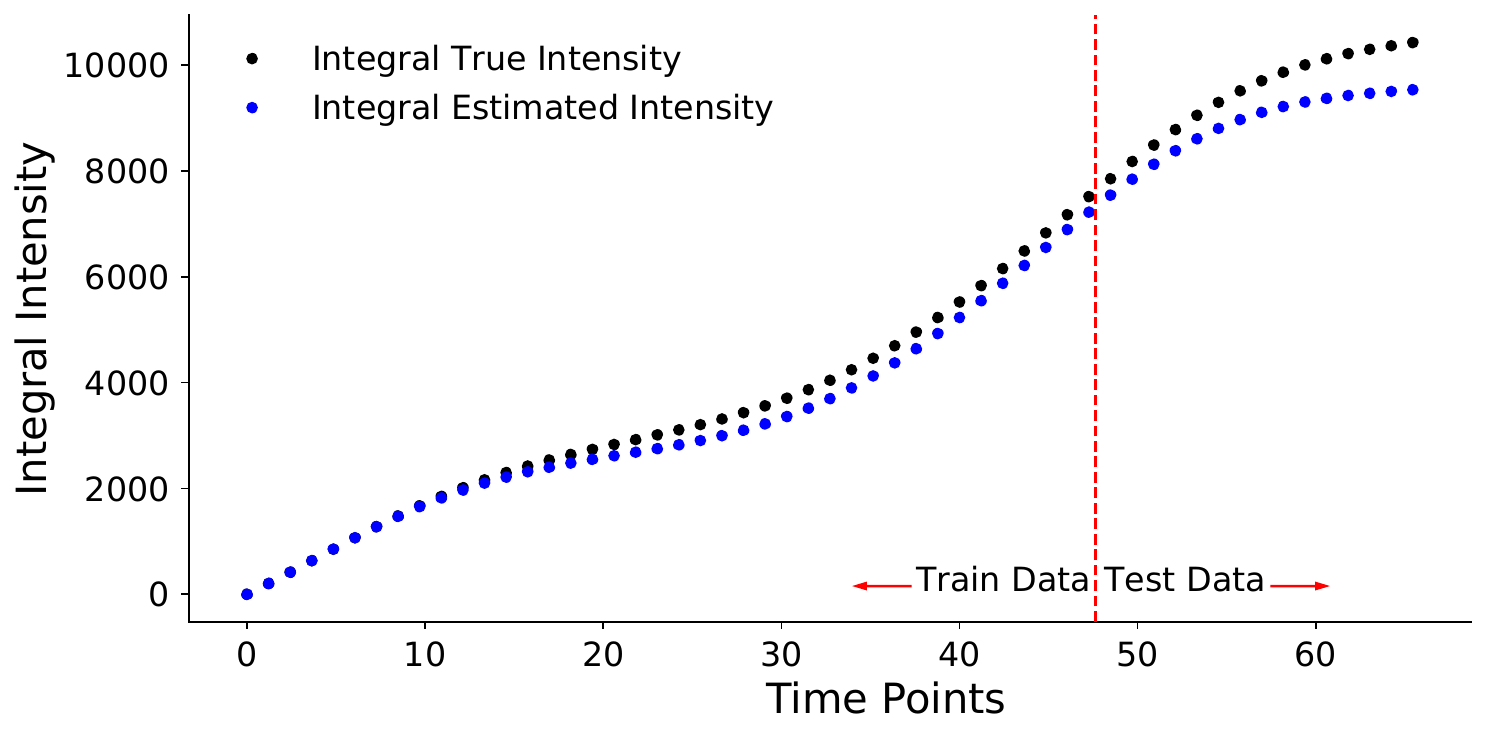}
  \caption{}
\end{subfigure} 

  \caption{Inferring the intensity of a log-Gaussian Cox Process. (a) compares the posterior distribution of the intensity estimated by \spiVAE to the true intensity function on train and  test data. (b) compares the posterior mean of the cumulative integral over time estimated by \spiVAE to the true cumulative integral on train and  test data. }
  \label{fig:1-d-lgcp}
\end{figure*}

Figure~\ref{fig:1-d-uncertainty}(a) presents results for \spiVAE with a cubic prior, Figure~\ref{fig:1-d-uncertainty}(b) with an RBF prior, and Figure~\ref{fig:1-d-uncertainty}(c) for standard Gaussian processes fitting using an RBF kernel. The mean absolute error (MAE) for all three methods are presented in Table~\ref{tbl:cubic}. Both, the mean estimates and the uncertainty from \spiVAE variants, are closer, and more constrained than the ones using Gaussian processes with HMC. Importantly, \spiVAE with cubic prior not only produces better point estimates but is able to capture better uncertainty bounds. We note that \spiVAE does not exactly replicate an RBF Gaussian process, but does retain the main qualitative features inherent to GPs - such as the concentration of the posterior where there is data. Despite \spiVAE ostensibly learning an RBF function class, differences are to be expected from the VAE low dimensional embedding. This simple example demonstrates that \spiVAE can be used to incorporate domain knowledge about the functions being modelled.

\begin{table}
\centering
\begin{tabular}{ll}
\toprule
Method         & Test MAE \\ \midrule
\spiVAE (cubic functions) & 10.47 \\ 
\spiVAE (Gaussian process with RBF kernel) & 33.15 \\ 
Gaussian process with RBF kernel & 67.37 \\
\bottomrule
\end{tabular}
\caption{Test results of fitting to a cubic function with noise $y \sim \mathcal{N}(x^3, 9)$.}
\label{tbl:cubic}
\end{table}

In many scenarios, learning just the mapping of inputs to outputs is not sufficient as other functional properties are required to perform useful (interesting) analysis. For example, using point processes requires knowing the underlying intensity function, however, to perform inference we need to calculate the integral of that intensity function too. Calculating this integral, even in known analytical form, is very expensive. Hence, in order to circumvent the issue, we use \spiVAE to learn both function values and its integral for the observed events. Figure \ref{fig:1-d-lgcp} shows \spiVAE prediction for both the intensity and integral of a simulated 1-D log-Gaussian Cox Process (LGCP).

In order to train \spiVAE to learn from the function space of 1-D LGCP functions, we first create a training set by drawing 10,000 different samples of the intensity function using an RBF kernel for 1-D LGCP. For each of the drawn intensity function, we choose an appropriate time horizon to sample $80$ observed events (locations) from the intensity function. \spiVAE is trained on the sampled $80$ locations with their corresponding intensity and the integral. \spiVAE therefore outputs both the instantaneous intensity and the integral of the intensity. The implementation details can be seen in the Appendix. For testing, we first draw a new intensity function (1-D LGCP) using the same mechanism used in training and sample $100$ events (locations).
As seen in Figure~\ref{fig:1-d-lgcp} our estimated intensity is very close to true intensity and even the estimated integral is close to the true integral. This example shows that the \spiVAE approach can be used to learn not only function evaluations but properties of functions.

\begin{figure*}[!tbp]
  \centering
\begin{subfigure}{.22\textwidth}
  \includegraphics[height=0.18\textheight]{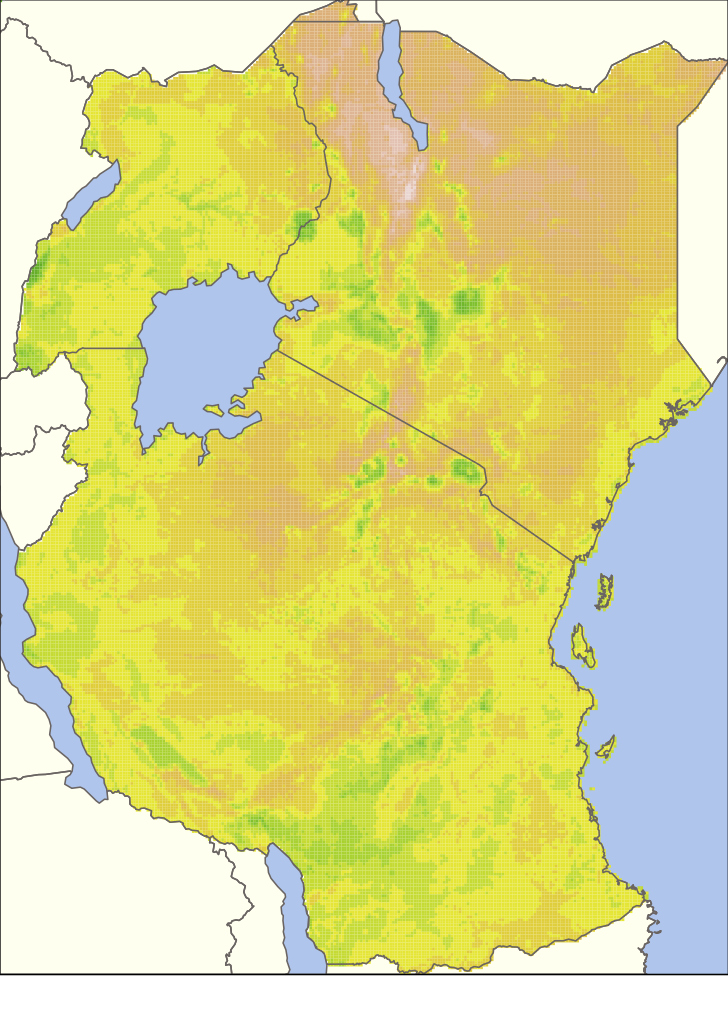}
 \caption{}
\end{subfigure}\quad
\begin{subfigure}{.22\textwidth}
  \includegraphics[height=0.18\textheight]{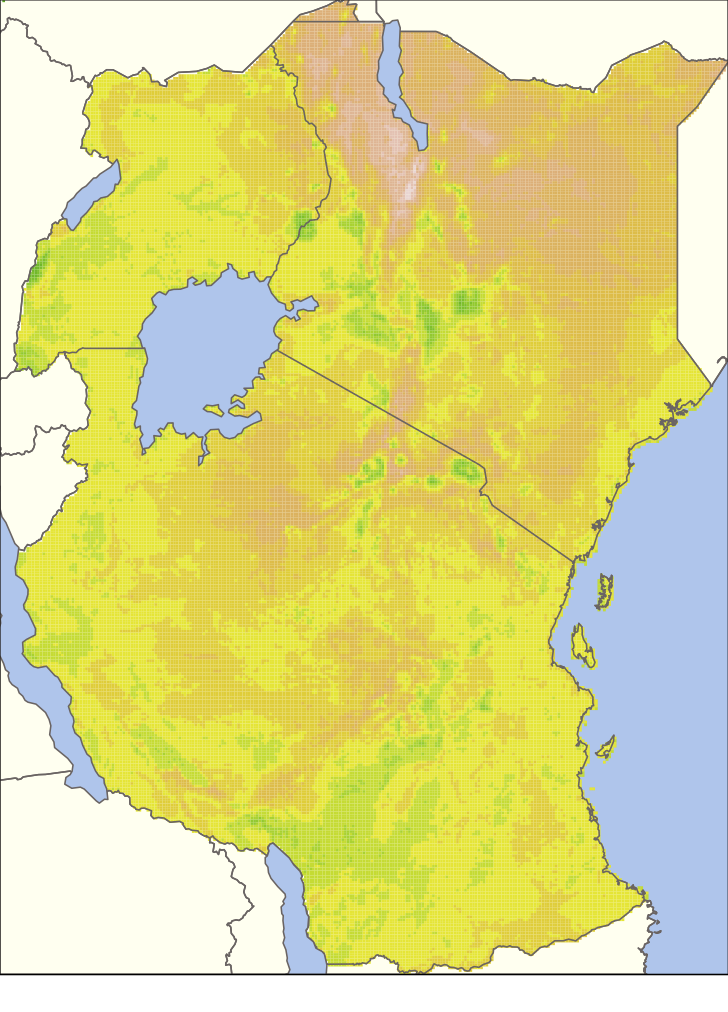}

 \caption{}
\end{subfigure} \quad
\begin{subfigure}{.22\textwidth}
  \includegraphics[height=0.18\textheight]{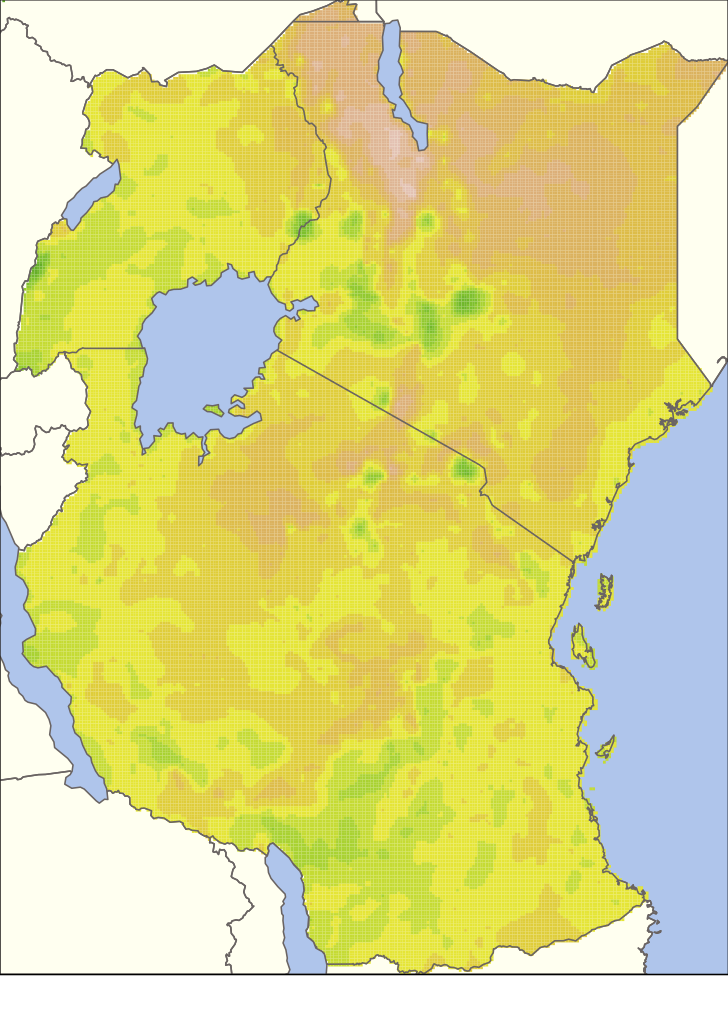}
 \caption{}
\end{subfigure} \quad
\begin{subfigure}{.22\textwidth}
  \includegraphics[height=0.18\textheight]{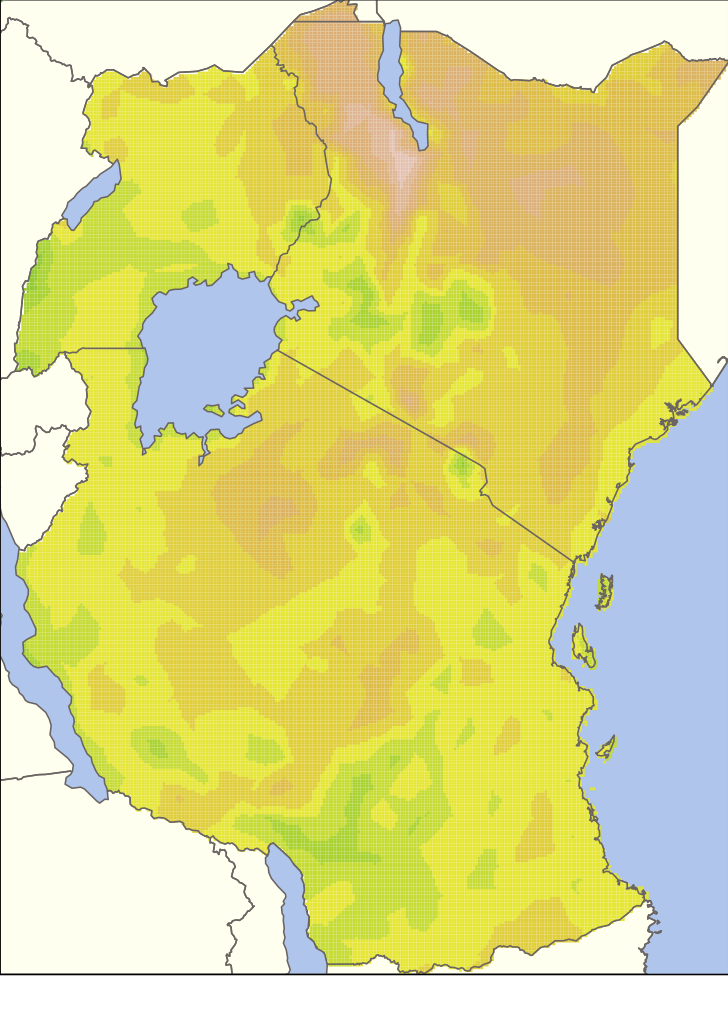}
 \caption{}
\end{subfigure} \quad
   \caption{
Deviation in land surface temperature for East Africa trained on 6000 random uniformly chosen locations  \citep{Ton2018SpatialFeatures}. Plots: (a) the data, (b)  our \spiVAE approach (testing MSE: 0.38), (c) a full rank GP with \Matern $\frac{3}{2}$ kernel (testing MSE: 2.47), and (d) a low rank SPDE approximation with 1046 basis functions \citep{Lindgren2011} and a \Matern $\frac{3}{2}$ kernel (testing MSE: 4.36). \spiVAE not only has substantially lower test error,  it captures fine scale features much better than  Gaussian processes or neural processes.}
  \label{fig:spatial}
\end{figure*}

\section{Results}
\label{sec:results}
\begin{figure*}[h]
  \begin{tabular}{ccc}
  10\% pixels & 20\% pixels & 30\% pixels \\
  \subfloat[Observation]{\includegraphics[width = 1.2in, height=1.2in]{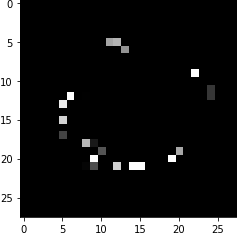}} &
  \subfloat[Observation]{\includegraphics[width = 1.2in, height=1.2in]{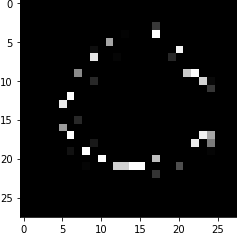}} &
  \subfloat[Observation]{\includegraphics[width = 1.2in, height=1.2in]{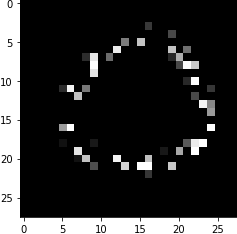}}\\
  \subfloat[Sample 1]{\includegraphics[width = 1.2in, height=1.2in]{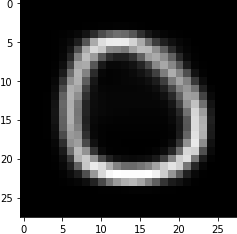}} &
  \subfloat[Sample 1]{\includegraphics[width = 1.2in, height=1.2in]{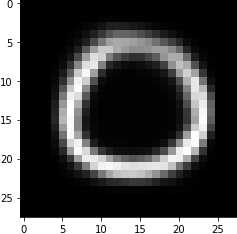}} &
  \subfloat[Sample 1]{\includegraphics[width = 1.2in, height=1.2in]{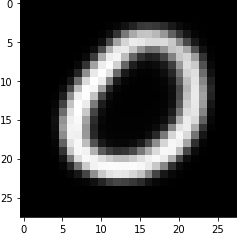}}\\
  \subfloat[Sample 2]{\includegraphics[width = 1.2in, height=1.2in]{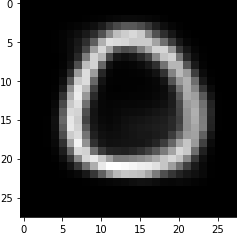}} &
  \subfloat[Sample 2]{\includegraphics[width = 1.2in, height=1.2in]{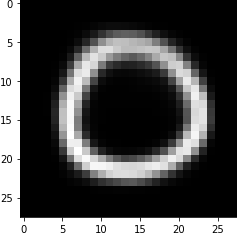}} &
  \subfloat[Sample 2]{\includegraphics[width = 1.2in, height=1.2in]{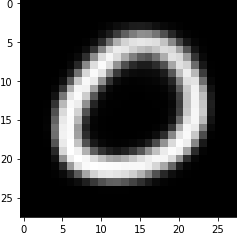}}\\
  \subfloat[Sample 3]{\includegraphics[width = 1.2in, height=1.2in]{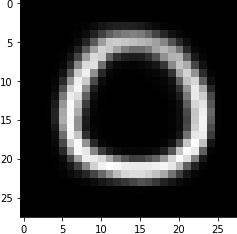}} &
  \subfloat[Sample 3]{\includegraphics[width = 1.2in, height=1.2in]{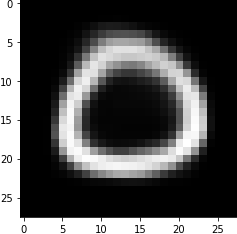}} &
  \subfloat[Sample 3]{\includegraphics[width = 1.2in, height=1.2in]{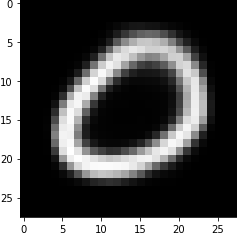}}
  \end{tabular}
  \caption{MNIST reconstruction after observing only 10, 20 or 30\% of pixels from original data.}
  \label{fig:MNIST}
\end{figure*}

Here we show applications of \spiVAE on three real world datasets. In our first example we use \spiVAE to predict the  deviation in land surface temperature in East Africa \citep{Ton2018SpatialFeatures}. We have the deviation in land surface temperatures for ${\sim}89{,}000$ locations across East Africa. Our training data consisted of 6,000 uniformly sampled locations. Temperature was predicted using only the spatial locations as inputs. Figure~\ref{fig:spatial} and Table~\ref{tbl:spatial} shows the results of the ground truth (a), our \spiVAE (b), a full rank Gaussian process with \Matern kernel \citep{Gardner2018} (c), and  low rank Gauss Markov random field (GMRF) (a widely used approach in the field of geostatistics) with $1,046$ ($\frac{1}{6}$th of the training size) basis functions \citep{Lindgren2011,Rue2009} (d). We train our \spiVAE model on $10^{7}$ functions draws from $2$-D GP with small lengthscales between $10^{-5}$ to $2$. $\Phi$ was set to be a \Matern layer ( see Appendix ~\ref{app:implementation}) with 1,000 centres followed by a two layer neural network of 100 hidden units in each layer. The latent dimension of \spiVAE was set to $20$. As seen in Figure~\ref{fig:spatial}, \spiVAE is able to capture small scale features and produces a far better reconstruction than the both full and low rank GP and despite having a much smaller latent dimension of $20$ vs 6,000 (full) vs 1,046 (low). The testing error for \spiVAE is substantially better than the full rank GP which leads to the question, why does \spiVAE perform so much better than a GP, despite being trained on samples from a GP? One possible reason is that the extra hidden layers in $\Phi$ create a much richer structure that could capture elements of non-stationarity \citep{Ton2018SpatialFeatures}. Alternatively, the ability to use state-of-the-art MCMC and estimate a reliable posterior expectation might create resilience to overfitting. The training/testing error for \spiVAE is $0.07/0.38$, while the full rank GP is $0.002/2.47$. Therefore the training error is 37 times smaller in the GP, but the testing error is only 6 times smaller in \spiVAE suggesting that, despite marginalisation, the GP is still overfitting. 

\begin{table}
\centering
\begin{tabular}{ll}
\toprule
Method         & Test MSE \\ \midrule
Full rank GP & 2.47 \\ 
\spiVAE    & 0.38 \\ 
low rank GMRF $(basis=1046)$    & 4.36  \\
\bottomrule
\end{tabular}
\caption{Test results for \spiVAE, ~a full rank GP, and low rank GMRF on \emph{land surface temperature for East Africa trained on 6000 random uniformly chosen locations  \citep{Ton2018SpatialFeatures}}.}
\label{tbl:spatial}
\end{table}

\begin{table}
\centering
\begin{tabular}{llr}
\toprule
Method         & RMSE  & NLL \\ \midrule
Full rank GP & 0.099 &  -0.258  \\ 
\spiVAE    & 0.112 & 0.006   \\ 
SGPR $(m=512)$    & 0.273 & 0.087   \\
SVGP $(m=1024)$  & 0.268 & 0.236   \\
\bottomrule
\end{tabular}
\caption{Test results for \spiVAE, ~a full rank GP and approximate algorithms SGPR and SVGP on \emph{Kin40K}.}
\label{tbl:kink}
\end{table}

Table~\ref{tbl:kink} compares \spiVAE on the \emph{Kin40K}~\citep{schwaighofer2003transductive} dataset to state-of-the-art full and approximate GPs, with results taken from \citep{wang2019exact}. The objective was to predict the distance of a robotic arm from the target given the position of all 8 links present on the robotic arm. In total we have 40,000 samples which are divided randomly into $\frac{2}{3}$ training samples and $\frac{1}{3}$ test samples. We train \spiVAE  on $N = 10^{7}$ functions drawn from an 8-D GP, observed at $K=200$ locations, where each of the $8$ dimensions had values drawn uniformly from the range $(-2,2)$ and lengthscale varied between $10^{-3}$ and $10$. Once \spiVAE was trained on the prior function we use it to infer the posterior distribution for the training examples in \emph{Kin40K}. Table~\ref{tbl:kink} shows results for  RMSE and negative log-likelihood (NLL)  of \spiVAE against various GP methods on test samples. The full rank GP results reported in~\citep{wang2019exact} are better than those from \spiVAE, but we are competitive, and far better than the approximate GP methods. We also note that the exact GP is estimated via maximising the log marginal likelihood in closed form, while \spiVAE performs full Bayesian inference; all posterior checks yielded excellent convergence measured via $\hat{R}$ and effective samples sizes. Calibration was checked using posterior predictive intervals. For visual diagnostics see the Appendix. 

Finally, we apply \spiVAE to the task of reconstructing MNIST digits using a subset of pixels from each image. Similar to the earlier temperature prediction task, image completion can also be seen as a regression task in 2-D. The regression task is to predict the intensity of pixels given the pixel locations. We first train  neural processes on full MNIST digits from the training split of the dataset, whereas \spiVAE is trained on $N = 10^{6}$ functions drawn from a 2-D GP.
The latent dimension of \spiVAE is set to be 40. As with previous examples, the decoder and encoder networks are made up of two layer neural networks. The hidden units for the encoder are 256 and 128 for the first and second layer respectively, and the reverse for decoder. 

Once we have trained \spiVAE we now use images from the test set for prediction. Images in the testing set are sampled in such a way that only 10, 20 or 30\% of pixel values are observed. Inference is performed with \spiVAE to predict the intensity at all other pixel locations using Eq.~\eqref{eqn:ppd}. As seen from Figure~\ref{fig:MNIST}, the performance of \spiVAE increases with increase in pixel locations available during prediction but still even with 10\% pixels our model is able to learn a decent approximation of the image. The uncertainty in prediction can be seen from the different samples produced by the model for the same data. As the number of given locations increases, the variance between samples decreases with quality of the image also increasing. Note that results from neural processes, as seen in Figure~\ref{fig:MNIST-NP}, look better than from \spiVAE. Neural processes performed better in the MNIST case because they were specifically trained on full MNIST digits from the training dataset, whereas piVAE was trained on the more general prior class of 2D GPs.

\section{Discussion and Conclusion}
\label{sec:discussion}

In this paper we have proposed a novel VAE formulation of a stochastic process, with the ability to learn function classes and properties of functions. Our \spiVAEs  typically have a small  (5-50) , uncorrelated latent dimension of parameters, so Bayesian inference with MCMC is straightforward and highly effective at successfully exploring the posterior distribution. This accurate estimation of uncertainty is essential in many areas such as medical decision-making.

\spiVAE combines the power of deep learning to create high capacity function classes, while ensuring tractable inference using fully Bayesian MCMC approaches. Our 1-D example in Figure \ref{fig:1-d-uncertainty} demonstrates that an exciting use of \spiVAE is to incorporate domain knowledge about the problem.  Monotonicity or complicated dynamics can be encoded directly into the prior \citep{Caterini2018}
on which \spiVAE is trained. Our log-Gaussian Cox Process example shows that not only functions can be modelled, but also properties of functions such as integrals.  Perhaps the most surprising result is the performance of \spiVAE on spatial interpolation. Despite being trained on samples from a Gaussian process, \spiVAE substantially outperforms a full rank GP. We conjecture this is due to the more complex structure of the feature
representation $\Phi$ and due to a resilience to overfitting. 

There are costs to using \spiVAE, especially the large upfront
cost in training. For complex priors, training could take days or weeks and will invariably require the heuristics and parameter searches inherent in applied deep learning to achieve a good performance.  However, once trained, a \spiVAE network is applicable on a wide range of problems, with the Bayesian inference MCMC step taking seconds or minutes. 

Future work should investigate the performance of \spiVAE on higher dimensional settings (input spaces $>10$). Other stochastic processes, such as Dirichlet processes, should also be considered.

\section*{Declarations}

\subsection*{ Funding}
 SB acknowledge the The Novo Nordisk Young Investigator Award (NNF20OC0059309) which also supports SM. SB also acknowledges the Danish National Research Foundation Chair grant, The Schmidt Polymath Award and The NIHR Health Protection Research Unit (HPRU) in Modelling and Health Economics. SM and SB acknowledge funding from the MRC Centre for Global Infectious Disease Analysis (reference MR/R015600/1) and Community Jameel. SF acknowledges the EPSRC (EP/V002910/2) and the Imperial College COVID-19 Research Fund.
\subsection*{ Conflict of interest/Competing interests}
NA
\subsection*{ Ethics approval }
NA
\subsection*{ Consent to participate}
NA
\subsection*{ Consent for publication}
NA
\subsection*{ Availability of data and materials}
All data used in the paper is available at \url{https://github.com/MLGlobalHealth/pi-vae}.
\subsection*{ Code availability }
Code is available at \url{https://github.com/MLGlobalHealth/pi-vae}.

\bibliography{references}
\clearpage
\onecolumn
\begin{appendices}
\section{MCMC diagnostics}
\label{app:mcmc}
\begin{figure*}[h]
    \includegraphics[width=\textwidth]{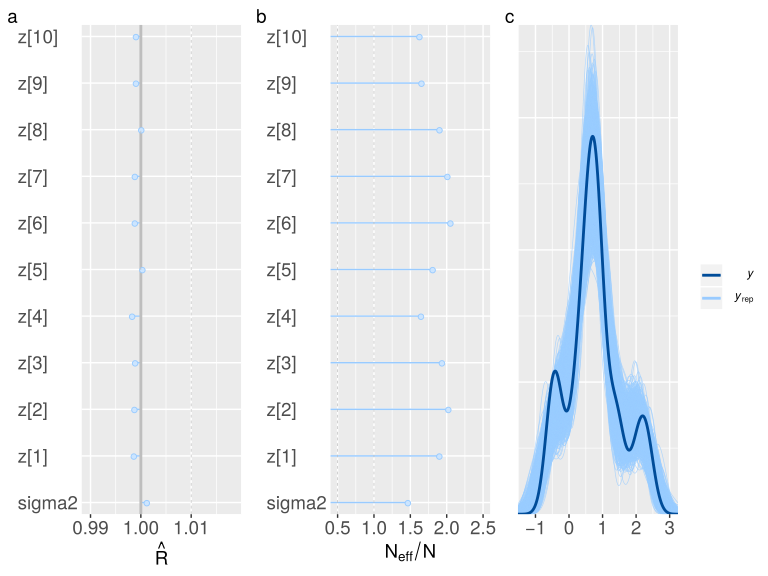}
    \caption{
  MCMC diagnostics for VAE inference presented in Figure~\ref{fig:1}: (a) and (b) shows the values for $\hat{R}$ and $\dfrac{N_{eff}}{N}$ for all parameters inferred with Stan. (c) shows the true distribution of observations along with the draws from the posterior predictive distribution.}
    \label{fig:appendix1}
\end{figure*}
\FloatBarrier
Figure~\ref{fig:appendix1} presents the MCMC diagnostics for the 1-D GP function learning example shown in Figure~\ref{fig:1}. Both $\hat{R}$ and effective sample size for all the inferred parameters (latent dimension of the VAE and noise in the observation) are well behaved with $\hat{R} \leq 1.01$ (Figure~\ref{fig:appendix1}(a)) and effective sample size greater than 1 (Figure~\ref{fig:appendix1}(b)). Furthermore, even the draws from the posterior predictive distribution very well capture the true distribution in observations as shown in Figure~\ref{fig:appendix1}(c).
\begin{figure*}[h]
    \includegraphics[width=\textwidth]{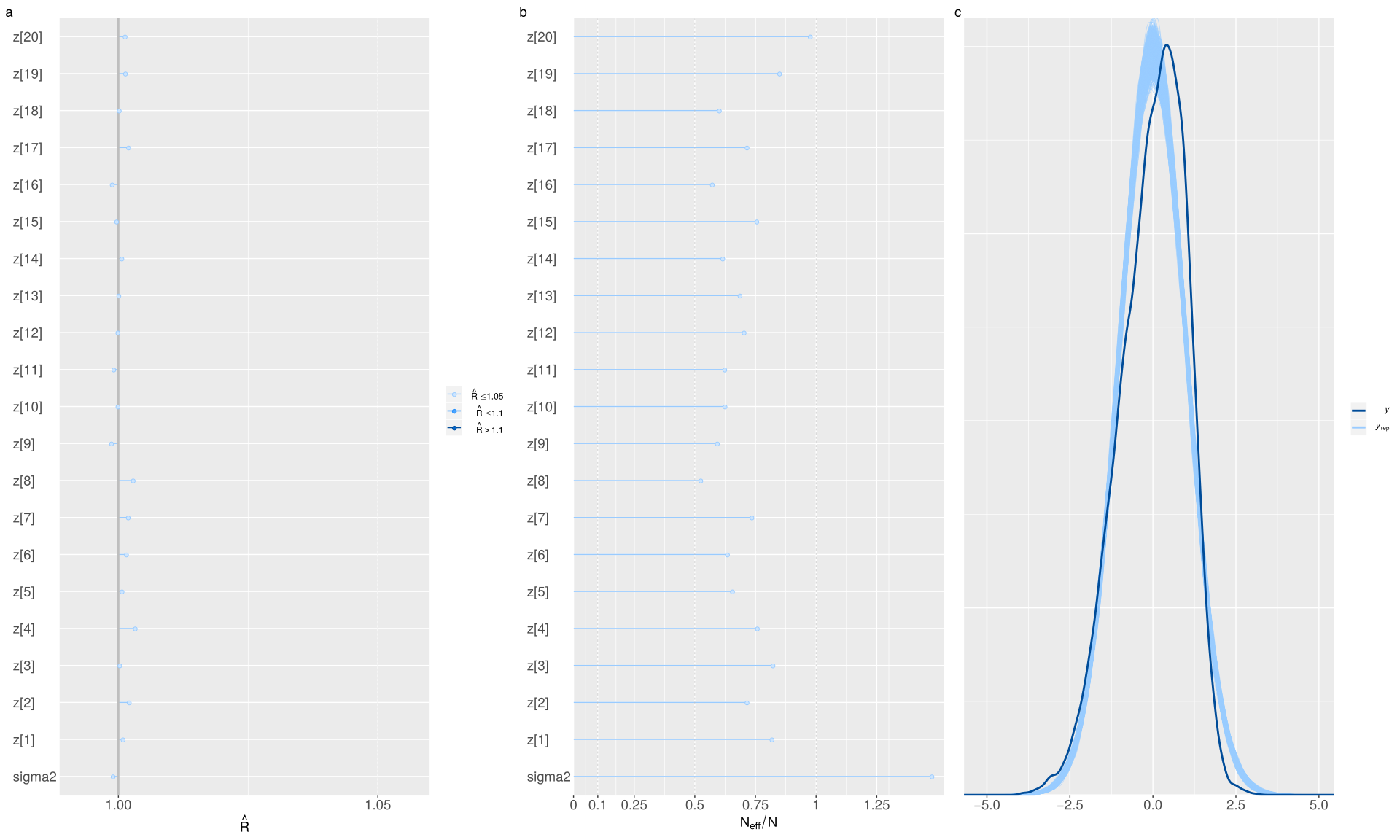}
    \caption{
  MCMC diagnostics for \spiVAE inference presented in Table~\ref{tbl:kink}: (a) and (b) shows the values for $\hat{R}$ and $\dfrac{N_{eff}}{N}$ for all parameters inferred with Stan. (c) shows the true distribution of observations along with the draws from the posterior predictive distribution.}
    \label{fig:appendix2}
\end{figure*}
\FloatBarrier

Figure~\ref{fig:appendix2} presents the MCMC diagnostics for the kin40K dataset with \spiVAE as shown in Table~\ref{tbl:kink}. Both $\hat{R}$ and effective sample size for all the inferred parameters (latent dimension of the VAE and noise in the observation) are well behaved with $\hat{R} \leq 1.01$ (Figure~\ref{fig:appendix2}(a)) and effective sample size greater than $0.5$ (Figure~\ref{fig:appendix2}(b)). Furthermore, the draws from the posterior predictive distribution are shown against the true distribution in observations as shown in Figure~\ref{fig:appendix2}(c).

\begin{figure*}[h]
    \includegraphics[width=\textwidth]{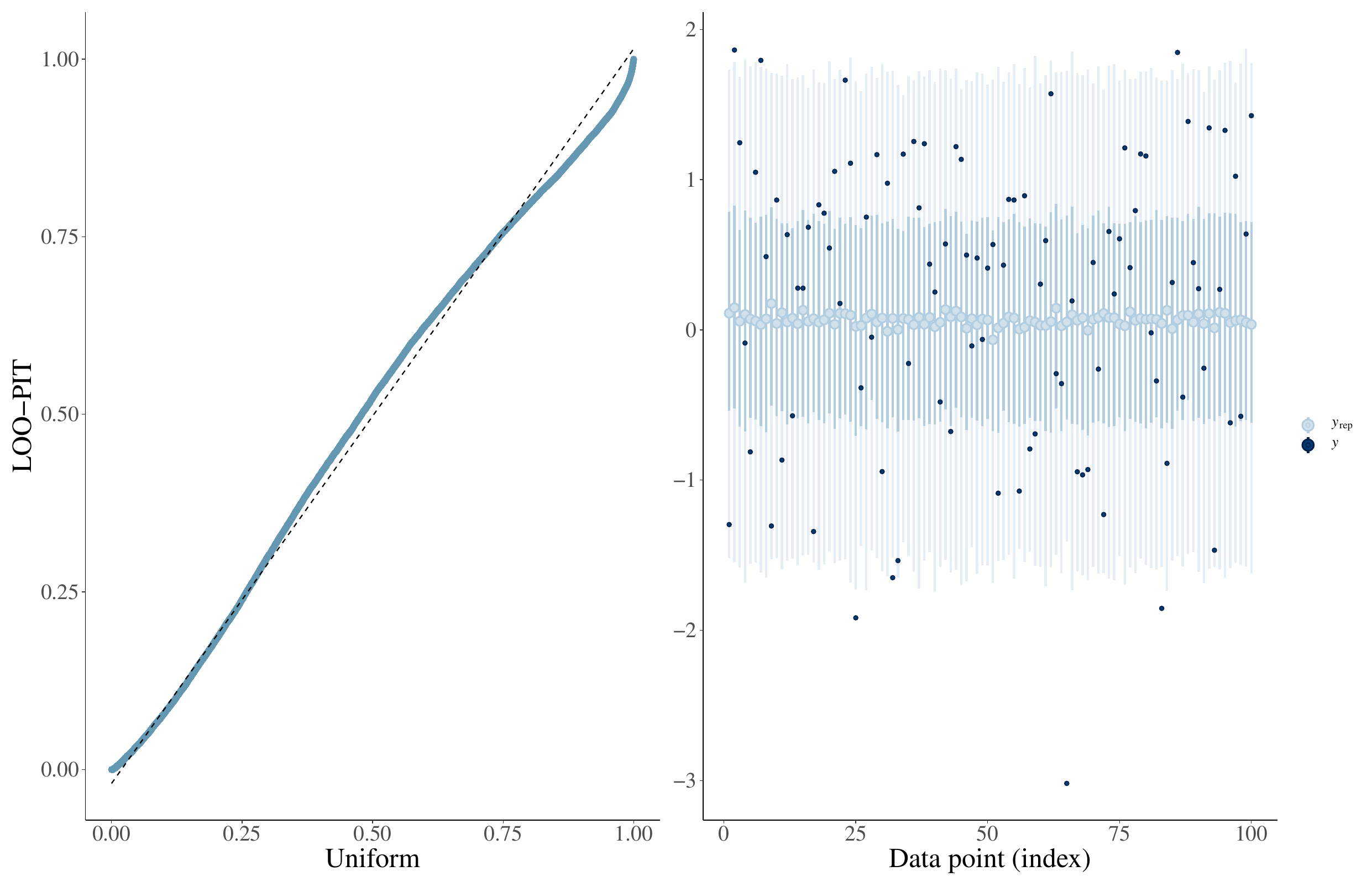}
    \caption{
  MCMC calibration for \spiVAE inference presented in Table~\ref{tbl:kink}: (a) shows the marginal predictive check using a leave one out probability integral transform and (b) shows the leave one out predictive intervals compared to observations.}
    \label{fig:appendix3}
\end{figure*}

Figure~\ref{fig:appendix3} presents the MCMC calibration plots for the posterior. Both the marginal predictive check and leave one out predictive intervals plots demonstrate that our posterior is well calibrated.

\FloatBarrier

\clearpage
\section{Algorithm}
\label{app:algo} 
\spiVAE proceeds in two stages. In the first stage (Algorithm \ref{algo:train}) we train \spiVAE, using a very large set of draws from a pre-specified prior class. In the second stage (Algorithm \ref{algo:infer}) we use the trained  \spiVAE from Algorithm \ref{algo:train} as a prior, combine this with data using a likelihood, and perform inference.  MCMC for Bayesian inference or optimization with a loss function are alternative approaches to learn the best  $\mathcal{Z}$ to explain the data. While these two algorithms are all that is needed to apply \spiVAE, for completeness Algorithm \ref{algo:sample} shows how one can use a trained \spiVAE, which encodes a stochastic process, to sample realisations from this stochastic process. 

    \begin{algorithm}[H]
        \caption{Prior Training for \spiVAE (stage 1)}  \label{algo:train}
        \begin{algorithmic}[1]
            \State Simulate draws from N functions evaluated at K points to create input data consisting of location, function value pairs: 
                        $\{(s_i^1,y_i^1),\ldots,(s_i^K,y_i^K)\}_{i=1}^{N}$ 

           \Repeat
           
           \For{each function i = 1, \ldots, N}
           \For{each location k = 1, \ldots, K}
            \State transform locations: $\Phi(s_i^k)$
            \State inner product with a linear basis:  
            \Statex \qquad \qquad \qquad $\hat{y}_{i,1}^k \leftarrow \beta_i^T\Phi(s_i^k)$
           \EndFor
           \State append $loss1$: $loss1$ $\leftarrow MSE(\hat{y}_{i,1}, y_i)$
           \State encode $\beta_i$ with VAE:  
           \Statex \qquad \qquad \qquad $\lbrack z_\mu,z_{sd}\rbrack^\top = e(\beta_i,\gamma)$
           \State reparameterize for $\mathcal{Z}$: $\mathcal{Z} \sim \mathcal{N}(z_\mu,z_{sd}^{2}\mathbb{I})$
           \State decode with VAE, $\hat{\beta}_i$ : $\hat{\beta}_i =  d(\mathcal{Z},\psi)$
           \For{each location k = 1, \ldots, K}
            \State transform locations: $\Phi(s_i^k)$
            \State inner product with decoded $\hat{\beta}_i$: 
            \Statex \qquad \qquad \qquad $\hat{y}_{i,2}^{k} = \hat{\beta}_i^{\top}\Phi(s_{i}^{k})$
           \EndFor
           \State append $loss2$: $loss2$ $\leftarrow MSE(\hat{y}_{i,2}, y_i)$
           \State minimize $loss1+loss2$ + $\text{KL}\left(\mathcal{Z} \| \mathcal{N}(0,\mathbb{I})\right)$ 
           \Statex \qquad \qquad \qquad to get $\Phi, \beta_i, \gamma, \psi$
           \EndFor
           \Until{termination criterion satisfied (epochs)}
        \end{algorithmic}
    \end{algorithm}
    \begin{algorithm}[H]
        \caption{Inference from \spiVAE (stage 2)} \label{algo:infer}
        \begin{algorithmic}[1] 
            \Require Trained decoder $d$ ($\psi$ fixed) and $\Phi$ (learnt from Algorithm~\ref{algo:train})
           \State \textbf{Input}: 
           $J$ observations consist of location, function value pairs: $\{(s_j,y_j)\}_{j=1}^J$.
           \State \textbf{Goal}: infer latent function with parameters  $\mathcal{Z}$.
           \State Sample $\mathcal{Z}$: $\mathcal{Z} \sim \mathcal{N}(0,\mathbb{I})$
            \State decode with VAE to get $\beta :=  d(\mathcal{Z},\psi)$
            \For{each location j}
            \State transform locations: $\Phi(s_j)$
            \State inner product with decoder $\beta$: $\hat{y}_j \leftarrow \beta^T\Phi(s_j)$
           \EndFor
           \State Perform Bayesian inference with MCMC for $\mathcal{Z}$ to obtain a set of draws from the posterior distribution:
           \begin{align*}
               p(\mathcal{Z}\mid d,y_1,s_1,\ldots,y_J,s_J,\Phi) & \propto p(y_1,\ldots,y_J\mid d,s_1,\ldots,s_J,\mathcal{Z},\Phi)p(\mathcal{Z}) \\
               & = p(y_1,\ldots,y_J\mid \hat y_1, \ldots, \hat y_J,\mathcal{Z})p(\mathcal{Z})
               \end{align*}
           \\ {\em (alternative to step 9:)} minimize an expected loss (e.g.~gradient decent with mean squared error): \Statex \qquad  $\argmin_{\mathcal{Z}} \sum_{j=1}^{J}\|\hat{y}_j$ - $y_{j}\|^2$. 
        \end{algorithmic}
    \end{algorithm}

\begin{algorithm}[H]
        \caption{Sampling from \spiVAE} \label{algo:sample}
        \begin{algorithmic}[1] 
            \Require Trained decoder $d$ ($\psi$ fixed) and $\Phi$ (learnt from Algorithm~\ref{algo:train})
            \State \textbf{Input}:
            Locations $s_1, \ldots, s_J$ where we want to evaluate the sampled function.
            \State Sample $\mathcal{Z}$: $\mathcal{Z} \sim \mathcal{N}(0, \mathbb{I})$
            \State decode with VAE to get : $\beta :=  d(\mathcal{Z},\psi)$
            \For{each location j}
            \State transform locations: $\Phi(s_j)$
            \State inner product with $\beta$: $\hat{y}_j \leftarrow \beta^T\Phi(s_j)$
           \EndFor
        \end{algorithmic}
    \end{algorithm}


\section{MNIST Example}
\begin{figure*}[tbp]
  \begin{tabular}{ccc}
  10\% pixels & 20\% pixels & 30\% pixels \\
  \subfloat[Observation]{\includegraphics[width = 1.5in, height=1.2in]{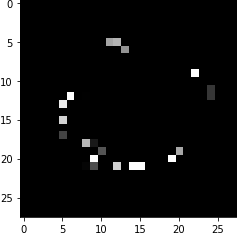}} &
  \subfloat[Observation]{\includegraphics[width = 1.5in, height=1.2in]{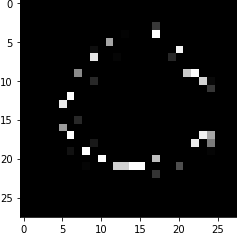}} &
  \subfloat[Observation]{\includegraphics[width = 1.5in, height=1.2in]{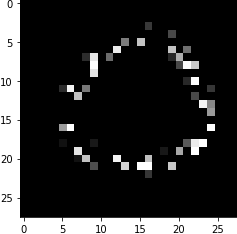}}\\
  \subfloat[Sample 1]{\includegraphics[width = 1.5in, height=1.2in]{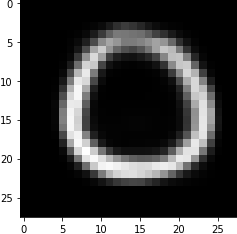}} &
  \subfloat[Sample 1]{\includegraphics[width = 1.5in, height=1.2in]{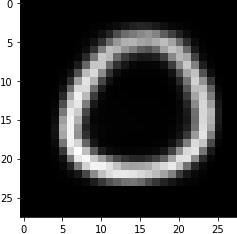}} &
  \subfloat[Sample 1]{\includegraphics[width = 1.5in, height=1.2in]{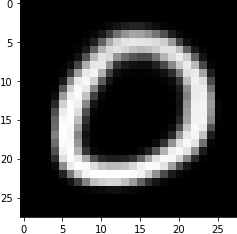}}\\
  \subfloat[Sample 2]{\includegraphics[width = 1.5in, height=1.2in]{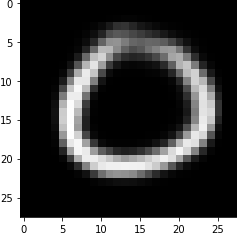}} &
  \subfloat[Sample 2]{\includegraphics[width = 1.5in, height=1.2in]{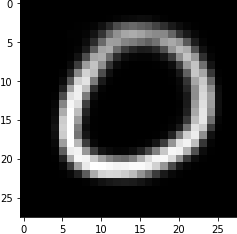}} &
  \subfloat[Sample 2]{\includegraphics[width = 1.5in, height=1.2in]{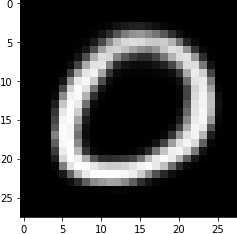}}\\
  \subfloat[Sample 3]{\includegraphics[width = 1.5in, height=1.2in]{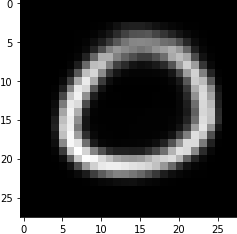}} &
  \subfloat[Sample 3]{\includegraphics[width = 1.5in, height=1.2in]{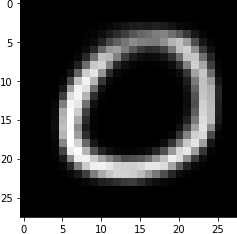}} &
  \subfloat[Sample 3]{\includegraphics[width = 1.5in, height=1.2in]{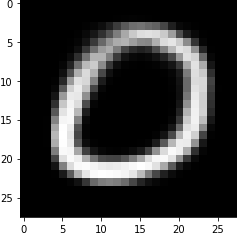}}
  \end{tabular}
  \caption{MNIST reconstruction using neural processes after observing only 10, 20 or 30\% of pixels from original data.}
  \label{fig:MNIST-NP}
\end{figure*}

Figure \ref{fig:MNIST-NP} below is the MNIST example referenced in the main text for neural processes.

\section{Implementation Details}
\label{app:implementation}
All  models were implemented with PyTorch~\citep{paszke2019pytorch} in Python. For Bayesian inference Stan~\citep{Carpenter2017a} was used. For training while using a fixed grid, when not mentioned in main text, in each dimension was on the range -1 to 1. Our experiments ran on a workstation with two NVIDIA GeForce RTX 2080 Ti cards.

\textbf{RBF and Matérn layers:} RBF and Matérn layers are implemented as a variant of the original RBF networks as described in~\citep{Broomhead1988Mar,Park1991Jun}. In our setting we define a set of $C$ trainable centers, which act as fixed points. Now for each input location we calculate a RBF or Matérn Kernel for all the fixed points. These calculated kernels are weighted for each fixed center and then summed over to create a scalar output for each location. We can describe the layer as follows:- 

\begin{equation*}
 \Phi(s) = \sum_{i=1}^{c} \alpha_{i} \rho\left(\lVert s-c_i \rVert\right)    
\end{equation*}

where $s$ is an input location (point), $\alpha_i$ is the weight for each center $c_i$, and $\rho\left(\lVert x-c_i \rVert\right)$ is RBF or Matérn kernel for RBF and Matérn layer respectively.

\subsection*{LGCP simulation example}

We study the function space of 1-D LGCP realisations. We define a nonnegative intensity function at any time $t$ as $\lambda(t): T  \rightarrow \mathcal{R}^+$. The number of events in an interval $[t_1,t_2]$ within some time period, $y_{[t_1,t_2]}$, is distributed Poisson via the following Bayesian hierarchical model:
\begin{align}
    Z(t)&\sim \text{GP}(0,k) \nonumber\\
    \lambda(t) &= \gamma \cdot \text{exp}(Z(t)) \nonumber\\
    y_{[t_1,t_2]} &\sim \text{Poisson}\left(\int_{t_1}^{t_2}\lambda(t) dt\right) 
     \label{eq:LGCP_hierarchy}
\end{align}
where $\gamma$ is a constant event rate, set to $5$ in our experiments.

To train \spiVAE on functions from an LGCP, we draw 10,000 samples from Eq~\eqref{eq:LGCP_hierarchy}, assuming  $k$ is an RBF kernel/covariance function with form $e^{-\sigma  x^2}$, with inverse lengthscale $\sigma$ chosen randomly from the set $\{8,16,32,64\}$. We then choose an observation window sufficiently large to ensure that $80$ events are observed. This approach is meant to simulate the situation in which we observe a point process until a certain number of events have occurrence, at which point we conduct inference~\citep{Mishra2016}. 

Given the set of  $80 \times 10,000$ events, we train \spiVAE with their corresponding intensity and integral of the intensity over the corresponding observation window. The integral is calculated numerically. We concatenate the integral of the intensity at the end with the intensity itself (value of the function evaluated the specific location). Note, in this setup we have $\beta$ setup as a $2-D$ vector, first value corresponding to the intensity and second to the integral of the intensity. The task for \spiVAE is to simultaneously learn both the instantaneous intensity and the integral of the intensity. At testing, we expand the number of events (and hence the time horizon) to $100$, and compare the intensity and integral of \spiVAE compared to the true LGCP. As seen in Figure~\ref{fig:1-d-lgcp}, in this extrapolation, our estimated intensity is very close to the true intensity and even the estimated integral is close to the true (numerically calculated) integral. This example shows that the \spiVAE approach can be used to learn not only function evaluations but properties of functions.

\end{appendices}

\end{document}